\documentclass[lettersize,journal]{IEEEtran}
\usepackage{amsmath,amsfonts}
\usepackage{algorithmic}
\usepackage{array}
\usepackage[caption=false,font=normalsize,labelfont=sf,textfont=sf]{subfig}
\usepackage{textcomp}
\usepackage{stfloats}
\usepackage{url}
\usepackage{verbatim}
\usepackage{graphicx}
\def\BibTeX{{\rm B\kern-.05em{\sc i\kern-.025em b}\kern-.08em
    T\kern-.1667em\lower.7ex\hbox{E}\kern-.125emX}}
\usepackage{balance}

\usepackage{bm}
\usepackage{color}
\usepackage{multirow}

\newcommand{\tabincell}[2]{\begin{tabular}{@{}#1@{}}#2\end{tabular}}
\usepackage{algorithmic}
\usepackage[ruled]{algorithm2e}
\usepackage{cite}

\usepackage{booktabs}
\DeclareMathOperator*{\argmax}{argmax}

\begin{document}
\title{CGIBNet: Bandwidth-constrained Communication \\with Graph Information Bottleneck in Multi-Agent Reinforcement Learning}
	\author{Qi Tian, Kun Kuang, Baoxiang Wang, Furui Liu, and Fei Wu, \IEEEmembership{Senior Member,~IEEE}
	\thanks{Qi Tian, Kun Kuang and Fei Wu are with College of Computer Science and Technology, Zhejiang University, Hangzhou 310027, China. (email: tianqics, kunkuang, wufei@zju.edu.cn).}
	\thanks{Baoxiang Wang is with School of Data Science, Chinese University of Hong Kong (Shenzhen), Shenzhen 518000, China. (email: bxiangwang@gmail.com).}
	\thanks{Furui Liu is with Huawei Noah’s Ark Lab, Huawei Technologies, Shenzhen 518000, China. (email: liufurui2@huawei.com).}
}

\markboth{Journal of \LaTeX\ Class Files,~Vol.~18, No.~9, September~2020}%
{How to Use the IEEEtran \LaTeX \ Templates}

\maketitle

\begin{abstract}
Communication is one of the core components for cooperative multi-agent reinforcement learning (MARL). The communication bandwidth, in many real applications, is always subject to certain constraints. 
To improve communication efficiency, in this article, we propose to simultaneously optimize whom to communicate with and what to communicate for each agent in MARL. By initiating the communication between agents with a directed complete graph, we propose a novel communication model, named Communicative Graph Information Bottleneck Network (CGIBNet), to simultaneously compress the graph structure and the node information with the graph information bottleneck principle. The graph structure compression is designed to cut the redundant edges for determining whom to communicate with. The node information compression aims to address the problem of what to communicate via learning compact node representations. 
Moreover, CGIBNet is the universal module for bandwidth-constrained communication, which can be applied to various training frameworks (\emph{i.e.}, policy-based and value-based MARL frameworks) and communication modes (\emph{i.e.}, single-round and multi-round communication).
Extensive experiments are conducted in Traffic Control and StarCraft II environments. The results indicate that our method can achieve better performance in bandwidth-constrained settings compared with state-of-the-art algorithms.
\end{abstract}

\begin{IEEEkeywords}
multi-agent reinforcement learning, graph information bottleneck, bandwidth-constrained communication, multi-agent system.
\end{IEEEkeywords}

\section{Introduction}
\IEEEPARstart{M}{Multi-agent} reinforcement learning (MARL) has shown its powerful ability to cope with many complex decision-making tasks, such as sensor networks \cite{zhang2019collaborative}, traffic networks \cite{wang2020large} and strategy games \cite{liu2020attentive,luo2020multiagent,jiang2020model}.
In these scenarios, the communication mechanism is regarded as a promising way to improve team collaboration, as multiple agents can exchange their local observations or corresponding embeddings with each other during the execution phase to make a better joint decision.
Therefore, a variety of works have been proposed in the field of multi-agent communication to improve the performance of MARL \cite{sukhbaatar2016learning,foerster2016learning,peng2017multiagent,das2019tarmac,jiang2020graph,pesce2020improving,rangwala2020learning,du2020learning,kim2021communication,ahilan2021correcting}.
\begin{figure}[htbp]
	\begin{center}
		\includegraphics[width=0.95\linewidth]{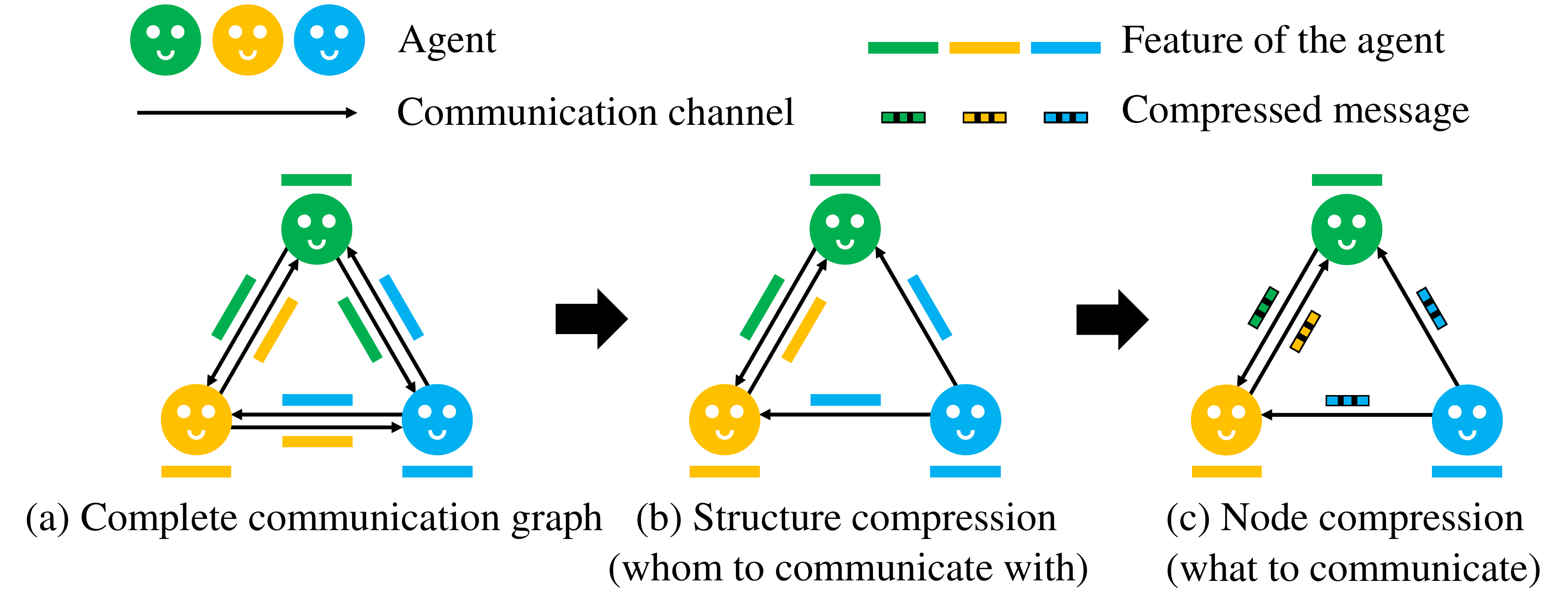}
		\caption{An illustration of graph structure and node information compression in our CGIBNet.
			(a) A directed complete graph is proposed to model multi-agent communication in MARL. (b) Graph structure compression is designed for identifying whom to communicate with. (c) Node information compression is used to learn what to communicate.}
		\label{intro}
	\end{center}
\end{figure}
In real multi-agent systems, however, communication resources are usually limited.
Under this setting, the key problem is how to efficiently exploit the available bandwidth and reduce the redundant communication among multi-agents. One possible solution is to optimize whom to communicate with and what to communicate for each agent in MARL. In this article, we focus on the problem of bandwidth-constrained communication in MARL, which is with the following challenges: (1) \textbf{Whom to communicate with?} Each agent needs to identify the agents that are necessary to communicate with for a good teamwork and establish an efficient cooperation protocol among multiple agents; (2) \textbf{What to communicate?} Each agent needs to deliver a concise and compact message, otherwise one large message may deplete the whole communication resources.

Recently, many methods \cite{liu2020multi,singh2018learning,kim2019learning,ETCNet,IMAC,NDQ} have been proposed to address the above challenges, but they have the following shortcomings:
1) Existing methods only focus on one of the challenges. \emph{e.g.}, \cite{liu2020multi,singh2018learning,kim2019learning,ETCNet} solve the first challenge, while \cite{IMAC,NDQ} solve the second. A seemingly straightforward idea is to combine these methods to simultaneously overcome both challenges. However, due to the lack of unified theoretical guidance and the compatibility problems of the methods, this combination can only achieve sub-optimal performance, as we will show in the experiments.
2) Many previous methods \cite{singh2018learning,kim2019learning,ETCNet,IMAC} need to learn an extra actor for message aggregation, hence are limited to policy-based MARL frameworks and cannot be applied to value-based MARL frameworks \cite{nguyen2020deep}.
3) All these methods are designed for single-round communication and cannot compress the communication bandwidth in multiple rounds of communication. Hence, it is of paramount importance to design a universal communication model that can simultaneously address the challenges of whom and what to communicate, and can be applied to various training frameworks (\emph{i.e.}, policy-based and value-based MARL frameworks) and communication modes (\emph{i.e.}, single-round and multi-round communication).

To achieve this goal, in this article, we propose a novel and universal multi-agent communication model named Communicative Graph Information Bottleneck Network (CGIBNet).
Specifically, we first model multi-agent communication with a directed complete graph as is shown in Fig. \ref{intro}(a). 
Then we derive two optimizable variational upper bounds based on the graph information bottleneck (GIB) principle and instantiate them into two regularizers to compress graph structure and node information. 
The graph structure compression regularizer is designed for addressing the challenge of whom to communicate with as is shown in Fig. \ref{intro}(b), by reducing the unnecessary communication channels between agents and maximally preserving information for decision-making.
Moreover, since we model multi-round communication as a hierarchical graph, \emph{i.e.}, multi-layer graph, this regularizer can also be used in each layer of the graph to reduce communication overhead in each communication round.
The node information compression regularizer is designed for solving the problem of what to communicate as is shown in Fig. \ref{intro}(c), by learning compact node representations to reduce bandwidth occupancy. 
Since the structure learning of CGIBNet is modeled as a link prediction task \cite{zhang2018link} instead of an actor-based scheduler, 
our model is learned implicitly with downstream reinforcement learning tasks in an end-to-end manner, hence, can be easily applied to the existing MARL frameworks (\emph{i.e.,} policy-based and value-based MARL) and communication models (\emph{i.e.}, single-round and multi-round communication).

We summarize our contributions as follows:

\begin{itemize}
	\item We investigate the problem of bandwidth-constrained communication in MARL by simultaneously considering with whom and what to communicate. We address this problem through the GIB principle and an optimization surrogate with rigorous theoretical guarantees.
	\item We propose a novel CGIBNet based on the GIB principle, to jointly compress the structure and node information of the communication graph for identifying with whom and what to communicate. CGIBNet is a universal module, which can be applied to various training frameworks (\emph{i.e.}, policy-based and value-based MARL frameworks) and communication modes (\emph{i.e.}, single-round and multi-round communication).
	\item Extensive experiments on both Traffic Control and StarCraft II datasets show that our proposed method outperforms the state-of-the-art algorithms in bandwidth-constrained communication settings.
\end{itemize}

	\begin{table*}[t]
	\setlength\tabcolsep{7.33pt}
	\centering
	\caption{Properties of different methods.}
	\label{related_work_table}
	\begin{tabular}{cccccccc}
		\toprule
		& \begin{tabular}[c]{@{}c@{}}Schedule structure\\ information \end{tabular}  & \begin{tabular}[c]{@{}c@{}}Value-based \\ framework \end{tabular} & \begin{tabular}[c]{@{}c@{}}Policy-based \\ framework \end{tabular} & \begin{tabular}[c]{@{}c@{}}Theoretical\\ guarantee \end{tabular}  & \begin{tabular}[c]{@{}c@{}}Compress structure\\ information \end{tabular}  & \begin{tabular}[c]{@{}c@{}}Compress node\\ information \end{tabular}& \begin{tabular}[c]{@{}c@{}}Multi-round\\ communication \end{tabular}  \\ \midrule
		IC3Net \cite{singh2018learning}      & $\checkmark$                              &                            & $\checkmark$                           &                       &                           &                                &                                                          \\ \specialrule{0em}{1pt}{1pt}
		G2A \cite{liu2020multi}      & $\checkmark$                              & $\checkmark$                          & $\checkmark$                           &                       &                           &                                &                                                          \\ \specialrule{0em}{1pt}{1pt}
		SchedNet \cite{kim2019learning} & $\checkmark$                              &                            & $\checkmark$                           &                       &$\checkmark$                           &                               &                                                          \\ \specialrule{0em}{1pt}{1pt}
		ETCNet \cite{ETCNet} & $\checkmark$                              &                            & $\checkmark$                           & $\checkmark$                      &$\checkmark$                           &                               &                                                          \\ \specialrule{0em}{1pt}{1pt}
		NDQ \cite{NDQ}      &                                & $\checkmark$                          & $\checkmark$                           & $\checkmark$                     &                          &$\checkmark$                                &                                                          \\ \specialrule{0em}{1pt}{1pt}
		IMAC \cite{IMAC}     & $\checkmark$                              &                            & $\checkmark$                           & $\checkmark$                     &                          &$\checkmark$                                &                                                          \\ \specialrule{0em}{1pt}{1pt}
		CGIBNet (ours)     & $\checkmark$                              & $\checkmark$                          & $\checkmark$                           & $\checkmark$                     & $\checkmark$                         & $\checkmark$                              & $\checkmark$                                                        \\ \bottomrule
	\end{tabular}
\end{table*}

\section{Related work} \label{related work}
\;\;\;\textbf{MARL communication.} 
Communication plays an important role in the MARL research since it can alleviate the non-stationary problem of multi-agent systems, thereby improving the group’s cooperation \cite{hernandez2019survey}.
Recently, a variety of works have been proposed to explore communication protocols in MARL, such as improving the message aggregation mechanism \cite{sukhbaatar2016learning,foerster2016learning,peng2017multiagent,das2019tarmac,jiang2020graph}, introducing memory modules \cite{pesce2020improving,rangwala2020learning}, learning world dynamic models \cite{du2020learning,kim2021communication}, adjusting experience buffers \cite{ahilan2021correcting}, \emph{etc}.
However, all the above methods require each agent to receive the messages from all agents in each control cycle to take an action, which is not suitable for practical scenarios where communication bandwidth is limited. 
To efficiently utilize finite communication resources, 
the existing works reduce the communication burden from two points: 1) who to communicate with (structure information) and 2) what to communicate (node information).

For the first point, researchers introduce various gating mechanisms to reduce communication channels between agents. 
\emph{e.g.,} IC3Net \cite{singh2018learning} models the gate by learning an extra actor, while G2A \cite{liu2020multi} achieves this goal through hard attention and combines with soft attention to improve its performance.
However, IC3Net and G2A do not explicitly constrain or penalize the communication frequency, thus these methods still occupy a lot of communication overhead.
SchedNet \cite{kim2019learning} and ETCNet \cite{ETCNet} leverage actor-based weight generators to explicitly restrict bandwidth occupancy, 
thus these methods can save some communication channels.
Unfortunately, all the existing methods have some application limitations. 
Specifically, IC3Net, SchedNet and ETCNet can only be applied to policy-based MARL frameworks. 
This is because they model communication scheduler as an actor $\mu$ and directly optimize the parameter $\theta$ of $\mu$ by policy gradient theorem \cite{sutton2000policy}, \emph{i.e.}, $A\nabla_{\theta}\log \mu$, which is not feasible in value-based frameworks since the policy gradient theorem is not applicable.
For a similar reason, these methods can only achieve single-round communication, because recursive optimization gradient $A\nabla_{\theta}(\log \mu(\nabla_{\theta}\log \mu))$ in multiple rounds of communication does not conform to the policy gradient theorem. 
G2A is not limited by the training frameworks, however, this method can only be used for single-round communication since its message aggregation module (\emph{i.e.}, BiLSTM \cite{huang2015bidirectional}) is not compatible with the multi-round communication process.

For the second point, researchers introduce various regularizers to learn compact message representations. \emph{e.g.,} NDQ \cite{NDQ} derives two regularizers based on mutual information to ensure the expressiveness and succinctness of the message. However, this method does not design a message aggregation module but uses fully connected communication, resulting in sub-optimal message representations and poor performance in some specific scenarios. \emph{e.g.}, target communication is needed \cite{das2019tarmac}.
IMAC \cite{IMAC} is most relevant to our work since it also introduces the information bottleneck (IB) principle \cite{alemi2017deep}.
We emphasize the differences between our method and IMAC as follows:
1) Theory: The IB principle used by IMAC can only solve the problem of what to communicate (\emph{i.e.}, compress message representations) for each agent, but has no theoretical guidance on who to communicate with (\emph{i.e.}, reduce communication channels). However, we extend the IB principle to the graph information bottleneck (GIB) principle, which can simultaneously solve the above two problems under a unified framework. 
2) Instantiation: Although IMAC claims that it can determine each agent is communicating to whom by learning a message aggregation module, we find this module cannot reduce communication channels between agents since it still uses the fully connected style. Therefore, no matter in theory or implementation, IMAC cannot reduce communication overhead in terms of who to communicate with. 
Furthermore, the message aggregation module of IMAC is modeled as an actor, which means that it has many usage restrictions similar to IC3Net, SchedNet and ETCNet.
Different from these methods, we model the message scheduler as a link prediction task \cite{zhang2018link} in graph neural networks, which can be naturally combined with our deduced information-theoretic regularizer and achieve end-to-end training without the above application limitations.

For better comparison, we list the properties of different methods as shown in Table \ref{related_work_table}. Note that the difference between `Schedule structure information' and `Compress structure information' is that the latter explicitly constrains or penalizes the redundant structural information.

\textbf{Information bottleneck.} Information bottleneck (IB), originally proposed for signal processing, tries to find a short code of the input signal but preserves maximum information of the code \cite{tishby1999information}. Alexander et al. \cite{alemi2017deep} first introduce the IB into deep neural networks, then the IB is applied to various domains \cite{chalk2016relevant,sinha2021dibs,goyal2019infobot}. Recently, 
Wu et al. \cite{wu2020graph} extend the IB theory to the graph data to learn compact node representations, which can improve model robustness against adversarial examples \cite{szegedy2014intriguing}. 
However, they assume that the graph data satisfies the local-dependence assumption and the underlying graph is a static graph (\emph{i.e.}, each layer of the graph structure is consistent and constant). Instead, we deal with a dynamic graph (\emph{i.e.}, the structure of the communication graph is different at each timestep). Meanwhile, we assume the structure of the lower layer is based on the upper layer in multiple rounds of communication. These lead to the differences between our method and their method in terms of both theoretical derivation and instantiation.

\section{Preliminary}
\subsection{Dec-POMDP and cooperative MARL.}
A fully cooperative multi-agent task can be described as a Decentralized Partially Observable Markov Decision Process (Dec-POMDP) \cite{oliehoek2016concise} and defined as a tuple $\langle \emph{N}, \mathcal{S}, \bm{\mathcal{A}}, \mathcal{T}, \mathcal{R}, \bm{\mathcal{O}}, \Omega, \gamma \rangle$, 
where $\emph{N}$ represents the number of agents.
$\mathcal{S}$ represents the true state space of the environment.
At each timestep $t \in \mathcal{Z}^{+}$, each agent $i \in \emph{N} \equiv \left\{ 1,\dots, n \right\}$ takes an action $a_i \in \mathcal{A}$, forming a joint action $\boldsymbol{a} \in \bm{\mathcal{A}} \equiv \mathcal{A}^n$.
Let $\mathcal{T}(s'|s, \boldsymbol{a}): \mathcal{S} \times \bm{\mathcal{A}} \rightarrow \mathcal{S}$ represents the state transition function.
All agents share the global reward function $r(s,\boldsymbol{a}):\mathcal{S} \times \bm{\mathcal{A}} \rightarrow \mathcal{R}$.
Since we consider a partially observable setting, each agent receives an individual observation $o_i \in \Omega$ according to the observation function $\mathcal{O}(s,a):\mathcal{S} \times \mathcal{A} \rightarrow \Omega$.
$\gamma \in [ 0,1 )$ represents the discount factor.
The current timestep superscript $t$ is omitted for all variables if there is no ambiguity.

\subsection{Policy-based and value-based MARL training frameworks.}
\emph{1) Policy-based MARL:} Independent actor-critic (IAC) \cite{foerster2018counterfactual} is the simplest multi-agent policy gradient method, which treats each agent as an independent learner. For each agent $i$, the main idea is to directly adjust the parameters $\theta_{i}$ of the actor policy $\pi_i$ in order to maximize the objective by taking steps in the direction of its gradient. By the policy gradient theorem \cite{sutton2000policy}, the gradient of the objective is
\begin{equation}
	\label{IAC}
	\nabla_{\theta_{i}} \mathcal{L}_{MARL,i}^{IAC} = \mathbb{E}_{\pi_i} [A_i (o_i, a_i) \nabla_{\theta_{i}} \log \pi_{i} (a_i|o_i)]\,,
\end{equation}
where $\mathbb{E}_{\pi_i}[\cdot]$ represents the sampled training data produced by the current policy $\pi_i$. $A_i (o_i, a_i)$ represents the advantage function. 
However, the decision-making of each agent is influenced by the non-stationarity problem \cite{gupta2017cooperative} in multi-agent systems, so the advantage function based on local information $o_i$ may be estimated inaccurately.
A reasonable alternative is the centralized advantage function $A_i(s, \boldsymbol{a})$ that incorporates global information $s$ and $\boldsymbol{a}$ to alleviate this problem.

Further, to reuse historical data, importance sampling is applied in \eqref{IAC}, as
\begin{equation}
	\nabla_{\theta_{i}} \mathcal{L}_{MARL,i} = \mathbb{E}_{\pi_{i,{\rm old}}} [\rho_i A_i(s, \boldsymbol{a}) \nabla_{\theta_{i}} \log \pi_{i} (a_i|o_i)]\,,
	\label{MAAC}
\end{equation}
where $\mathbb{E}_{\pi_{i,{\rm old}}}[\cdot]$ represents the sampled training data produced by the old policy $\pi_{i,{\rm old}}$ and $\rho_i=\pi_i(o_i|a_i)/\pi_{i,{\rm old}}(o_i|a_i)$ represents the importance weight to correct the bias.
Unfortunately, \eqref{MAAC} would lead to an excessively large policy update if without constraints. To penalize changes of policy and ensure the efficiency of importance sampling, Yu et al. \cite{yu2021surprising} propose a multi-agent version of the clipped PPO \cite{schulman2017proximal}
\begin{equation}
	\begin{aligned}
		\nabla_{\theta_{i}} \mathcal{L}_{MARL,i}^{MAPPO} & = \mathbb{E}_{\pi_{i,{\rm old}}} [\min(\zeta_{1},\zeta_{2}) \nabla_{\theta_{i}} \log \pi_{i} (a_i|o_i)] \\
		\zeta_{1} &= \rho_{i}A_{i}(s, \boldsymbol{a}) \\
		\zeta_{2} &= {\rm clip}(\rho_{i},1-\epsilon,1+\epsilon)A_{i}(s)\,,
	\end{aligned}
	\label{MAPPO}
\end{equation}
where $\epsilon$ is a hyperparameter to control the feasible range of the policy update. In this article, MAPPO will be used as one of the basic MARL frameworks for multi-agent communication.

\emph{2) Value-based MARL:} Value decomposition is a popular technical line in the value-based MARL \cite{hernandez2019survey}, among which QMIX \cite{rashid2018qmix} is a representative work.
It assumes that the total action-value function $Q_{tot}$ can be decomposed into individual action-value functions $Q_i$, which satisfies the following relation
\begin{equation}
	\argmax_{\bm{a}}Q_{tot}(s, \bm{a}) = 
	\begin{pmatrix}
		\argmax_{a_1}Q_1(o_1, a_1)   \\
		\vdots \\
		\argmax_{a_n}Q_n(o_n, a_n) \\
	\end{pmatrix}.
	\label{argmax}
\end{equation}
Further, the authors observe this representation can be generalized to the larger family of monotonic functions that are also sufficient but not necessary to satisfy \eqref{argmax}. Monotonicity can be enforced through the constraint
\begin{equation}
	\frac{\partial Q_{tot}}{\partial Q_i}  \geq 0,~ \forall i \in N.
	\label{monotonicity}
\end{equation}
QMIX uses a mixer hypernetwork \cite{ha2016hypernetworks} to implement this constraint. Then all network parameters can be optimized by a $Q$-learning style \cite{mnih2013playing} loss function as follows
\begin{equation}
	\mathcal{L}_{MARL}^{QMIX}=(y_{tot} - Q_{tot}(s, o_1, \dots, o_n, a_1, \dots, a_n) )^2,
	\label{QMIX}
\end{equation} 
where $y_{tot} = r+\gamma\max_{\bm{a}'} Q_{tot}^{-}(s^{\prime}, o_1^{\prime}, \dots, a_1^{\prime}, \dots)$ and 
$(\cdot)^{-}$ represents target networks.  $(\cdot)^{\prime}$ represents the variable at the next timestep. In this article, we choose QMIX as one MARL framework for the evaluation of communication modules.

\subsection{MARL with communication mechanism.}
In communication settings, multiple agents can transmit messages to each other for better collaboration.
Therefore, the policy $\pi_i$ in \eqref{MAPPO} and the individual action-value function $Q_i$ in \eqref{QMIX} can be rewritten as
\begin{equation}
	\begin{aligned}
		\text{Policy-based MARL (MAPPO):} &\quad \pi_{i} (a_i|o_i,\bm{m}_{-i}) \\
		\text{Value-based MARL (QMIX):} &\quad Q_i(o_i, a_i, \bm{m}_{-i}) \,,
	\end{aligned}
	\label{message}
\end{equation}
where $\bm{m}_{-i}=[m_1,\dots,m_{i-1},m_{i+1},\dots,m_N]$ represents the 	total messages that agent $i$ receives from its teammates.
Each individual message $m_j$ $(j\in {1,\dots,N}$ and $j\neq i)$ is usually obtained by dimensionality reduction mapping of local observations $m_j=\Psi(o_j)$, where $\Psi$ is usually implemented as a deep neural network.
In this article, we focus on bandwidth-constrained settings, which means that each agent $i$ cannot obtain all messages from other agents and can only pick some important ones.
\begin{figure}[t]
	\begin{center}
		\includegraphics[width=0.99\linewidth]{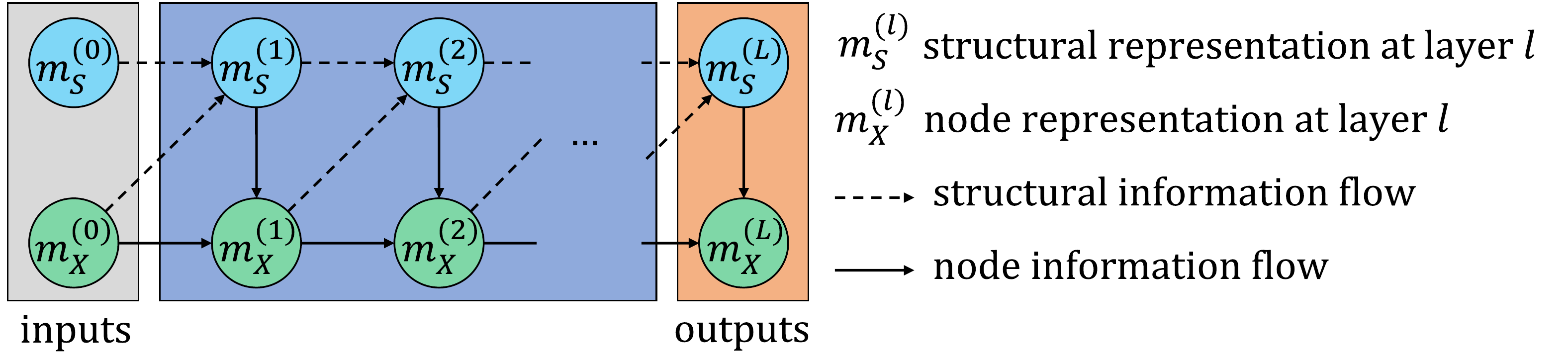}
		\caption{Information flow in the communication graph}
		\label{info_flow}
	\end{center}
\end{figure}
	\subsection{Deep variational information bottleneck.}
	Alexander et al. \cite{alemi2017deep} first propose the deep information bottleneck (IB) principle, given input $X$ and output $Y$, the goal is to learn a compressed representation $Z$ to maximize the mutual information between $Z$ and $Y$, while minimizing the mutual information between $Z$ and $X$ to learn a latent representation $Z$ that sufficiently forgets the spurious correlations that may exist in the input $X$ (sufficiency criterion), while still being predictive of $Y$ (minimum criterion). Formally,
	\begin{equation}
		\label{DIB_1}
		\max I\left(Z;Y\right) \,\,{\rm s.t.}\, I\left(X;Z\right)\le I_c\,,
	\end{equation}
	where $I\left(\cdot;\cdot\right)$ represents mutual information and $I_c$ represents an information constraint. To optimize this objective, Alexander et al. \cite{alemi2017deep} relax it into the alternative
	\begin{equation}
		\label{DIB_2}
		\max I\left(Z;Y\right)-\beta I\left(X;Z\right)\,,
	\end{equation}
	where $\beta$ is a predefined hyperparameter, which controls the tradeoff between compressibility and predictability. Inspired by this IB principle, a lot of variants have been proposed to solve various deep learning tasks (including computer vision \cite{chalk2016relevant,sinha2021dibs} and reinforcement learning \cite{goyal2019infobot,IMAC}). Our work is also based on this principle, but we extend it to the graph information bottleneck principle to build an efficient multi-agent communication module.

\section{Methodology}
In this section, we first illustrate the problem definition in this article. Then we detail the GIB principle in MARL communication to show its mechanism of saving communication resources.
Based on the GIB principle, we propose a novel communication model (CGIBNet).
Finally, we introduce the method to plug CGIBNet into both policy-based and value-based frameworks.

\subsection{Problem definition}
Our work focuses on how to learn a communication model to determine what messages to send and whom to address them in bandwidth-restricted settings.
In order to simultaneously solve these two problems under a unified framework,
we first model multi-agent communication as a graph, where nodes represent agents and edges represent communication channels between agents.
Therefore, the communication process can be regarded as the message passing \cite{gilmer2017neural} of each node information under the current graph structure.
Without loss of generality, we assume that there are $L$ rounds of communication between agents \cite{das2019tarmac} so that the communication graph has $L$ layers.
Then the structural representation $m_{S}^{(l)}$ and the node representation $m_{X}^{(l)}$ at the $l$-th layer can be denoted as
\begin{equation} 
	\begin{split}
		m_{S}^{(l)}=&\{m_{S,e}^{(l)}, m_{S,e}^{(l)}\in \{0,1\} \;{\rm and}\; e\in \mathcal{E}^{(l)}\} \\
		m_{X}^{(l)}=&\{m_{X,v}^{(l)}, m_{X,v}^{(l)}\in \mathbb{R}^{d} \;{\rm and}\; v\in \mathcal{V}^{(l)}\} \,,
	\end{split}
\end{equation}
where $\mathcal{E}^{(l)}$ and $\mathcal{V}^{(l)}$ indicate the valid edge set and the node set at the $l$-th layer, 0 and 1 indicate the existence of the corresponding edge, $d$ indicates the dimension of the node representation or the number of bits of each node message \cite{NDQ}.
Under the above definition, reducing the bandwidth of multi-agent communication is equivalent to compressing the flow of structural information and node information in the communication graph.
The former leads to fewer edge connections and the latter means that fewer bits can be used to express the whole node information. 
\subsection{GIB principle in MARL communication}
Inspired by the information bottleneck (IB) principle \cite{alemi2017deep}, which is an information-theoretic principle that extracts the minimal sufficient representation for the target task, we extend it to the graph information bottleneck (GIB) principle in multi-agent communication.
Specifically, we introduce two GIB-based criteria to compress the information flow over graph structures.

The first (minimum) criterion is to respectively constrain the structure-level mutual information $I(m_{S}^{(l-1)};m_{S}^{(l)})$ and the node-level mutual information $I(m_{X}^{(l-1)};m_{X}^{(l)})$ between adjacent layers to $I_{S}$ and $I_{X}$. This can obtain concise and compact information flow to reduce communication consumption.
In detail, different from \cite{wu2020graph}, they assume that each layer of the graph structure is consistent and constant, while we consider that the lower structure representation is related to the upper structure and node representation as is shown in Fig. \ref{info_flow}.
Thus we rewrite the structure-level and node-level mutual information and denote them as structural information bottleneck $SIB^{(l)}$ and node information bottleneck $XIB^{(l)}$, respectively:
\begin{equation}
	\label{SIB_XIB}
	\begin{split}
		SIB^{(l)}&=I(m_{S}^{(l-1)},m_{X}^{(l-1)};m_S^{(l)}) < I_{S}\\
		XIB^{(l)}&=I(m_X^{(l-1)},m_{S}^{(l)};m_X^{(l)})< I_{X}\,.
	\end{split}
\end{equation}

The second (sufficiency) criterion is to maximize the mutual information between all graph representations and the MARL target to preserve task-related information.
Since the communication model is jointly trained with other models, it can be achieved by minimizing the MARL task loss.

According to the above analysis, we formulate the GIB principle in bandwidth-limited communication tasks as
\begin{equation}
	\label{CGIB}
	\begin{split}
		\min\;& \mathcal{L}_{MARL} \\
		{\rm s.t.}\ SIB^{(l)}< I_{S}\; {\rm and}\; &XIB^{(l)}<I_{X},\; \forall \;l\in L\,,
	\end{split}
\end{equation}
where $\mathcal{L}_{MARL}$ represents the target task loss, and it can be implemented as $\mathcal{L}_{MARL}^{MAPPO}$ or $\mathcal{L}_{MARL}^{QMIX}$.
$I_{S}$ and $I_{X}$ limit the amount of information for structural representations and node representations to decrease bandwidth usage.

Since \eqref{CGIB} is hard to solve, similar to the approximation method in the information bottleneck community \cite{alemi2017deep,sinha2021dibs,goyal2019infobot,zhang2018link,IMAC} (\emph{i.e.}, \eqref{DIB_1} and \eqref{DIB_2}), we relax it as an unconstrained optimization problem
\begin{equation}
	\label{lag_mp_GIB}
	\min\; \mathcal{L}_{MARL}+\beta_{S}\sum_{l=1}^{L}SIB^{(l)}+\beta_{X}\sum_{l=1}^{L}XIB^{(l)}\,,
\end{equation}
where $\beta_{S}$ and $\beta_{X}$ represent two predefined hyperparameters that control the compression strength in the optimization. It is notoriously hard to directly optimize $SIB^{(l)}$ and $XIB^{(l)}$ due to the intractability of the mutual information terms. 
Thus we first derive the variational upper bound of $SIB^{(l)}$ as follows
\begin{equation}
	\label{eq1}
	\begin{split}
		&\,\,\,\,\,\, SIB^{(l)}=I(m_{S}^{(l-1)},m_{X}^{(l-1)};m_S^{(l)})\\
		&=\mathbb{E}_{p\left(m_{S}^{(l-1)},m_{X}^{(l-1)},m_S^{(l)}\right)}\log\left( \frac{p\left(m_{S}^{(l-1)},m_{X}^{(l-1)},m_S^{(l)}\right)}{p\left(m_{S}^{(l-1)},m_{X}^{(l-1)}\right)p\left(m_S^{(l)}\right)}\right)\\
		&=\mathbb{E}_{p\left(m_{S}^{(l-1)},m_{X}^{(l-1)},m_S^{(l)}\right)}\log\left( \frac{p\left(m_S^{(l)}|m_{S}^{(l-1)},m_{X}^{(l-1)}\right)}{p\left(m_S^{(l)}\right)}\right)\\
		&=\mathbb{E}_{p\left(m_{S}^{(l-1)},m_{X}^{(l-1)},m_S^{(l)}\right)}\log
		p\left(m_S^{(l)}|m_{S}^{(l-1)},m_{X}^{(l-1)}\right)\\
		&\,\,\,\,\, -\mathbb{E}_{ p\left(m_{S}^{(l-1)},m_{X}^{(l-1)},m_S^{(l)}\right)}\log
		p\left(m_S^{(l)}\right)\\
		&=\mathbb{E}_{p\left(m_{S}^{(l-1)},m_{X}^{(l-1)},m_S^{(l)}\right)}\log
		p\left(m_S^{(l)}|m_{S}^{(l-1)},m_{X}^{(l-1)}\right)\\
		&\,\,\,\,\, -\mathbb{E}_{p\left(m_S^{(l)}\right)}\log
		p\left(m_S^{(l)}\right)\,,\\
	\end{split}
\end{equation}
where the subscript $p(m_{S}^{(l-1)},m_{X}^{(l-1)},m_S^{(l)})$ of $\mathbb{E}$ is an abbreviation for $m_{S}^{(l-1)},m_{X}^{(l-1)},m_S^{(l)}\sim p(m_{S}^{(l-1)},m_{X}^{(l-1)},m_S^{(l)})$ and the subscript $p(m_S^{(l)})$ of $\mathbb{E}$ is an abbreviation for $m_S^{(l)}\sim p(m_S^{(l)})$. To save space, the following equations inherit this expression unless otherwise stated.
However, it is difficult to compute the marginal distribution $p\left(m_S^{(l)}\right)$, so let $r\left(m_S^{(l)}\right)$ be a variational approximation to this marginal. Then we have
\begin{equation} 
	\label{eq2}
	\begin{gathered}
		D_{KL}\left( p\left(m_S^{(l)}\right)||r\left(m_S^{(l)}\right)\right)\ge \, 0\\
		-\mathbb{E}_{p\left(m_S^{(l)}\right)}\log
		p\left(m_S^{(l)}\right) \le 
		-\mathbb{E}_{p\left(m_S^{(l)}\right)}\log
		r\left(m_S^{(l)}\right)\,,
	\end{gathered}
\end{equation}
where $D_{KL}(\cdot||\cdot)$ represents Kullback-Leibler (KL) divergence \cite{csiszar1975divergence}.
Substituting \eqref{eq2} into \eqref{eq1}, we have
\begin{figure*}[t]
	\begin{center}
		\includegraphics[width=0.99\linewidth]{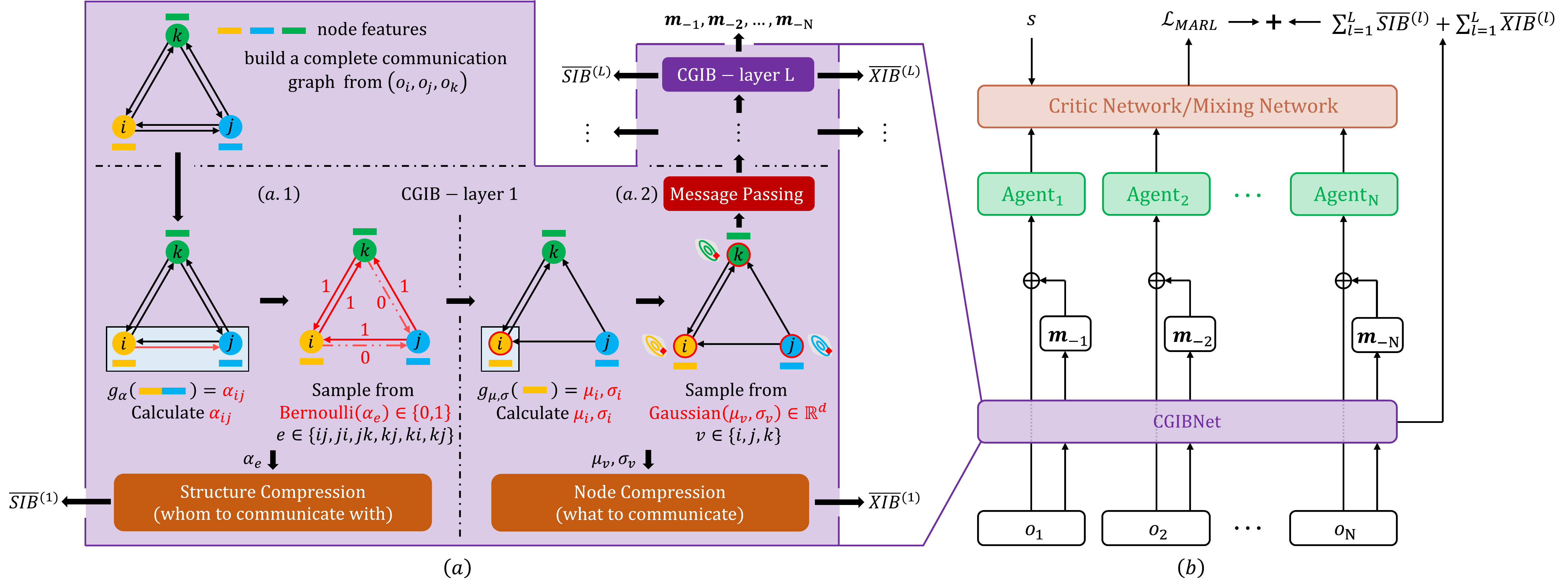}
		\caption{Overall framework of our method. (a) CGIBNet, where (a.1) represents structure compression and (a.2) represents node compression. (b) overall training pipeline.}
		\label{overall}
	\end{center}
\end{figure*}
\begin{equation}
	\begin{split}
		&\,\,\,\,\,\,\,\,\, SIB^{(l)}\\
		&\, \le \, \mathbb{E}_{ p\left(m_{S}^{(l-1)},m_{X}^{(l-1)},m_S^{(l)}\right)}\log
		p\left(m_S^{(l)}|m_{S}^{(l-1)},m_{X}^{(l-1)}\right)\\
		&\,\,\,\,\,\,\,\, - \mathbb{E}_{p\left(m_S^{(l)}\right)}\log
		r\left(m_S^{(l)}\right)\\
		&\le \, \mathbb{E}_{ p\left(m_{S}^{(l-1)},m_{X}^{(l-1)},m_S^{(l)}\right)}\log\left( \frac{p\left(m_S^{(l)}|m_{S}^{(l-1)},m_{X}^{(l-1)}\right)}{r\left(m_S^{(l)}\right)}\right)\,.
	\end{split}
\end{equation}
Thus the upper bound of $SIB^{(l)}$ can be written as
\begin{equation}
	\label{bound_SIB}
	\begin{split}
		&\widehat{SIB}^{(l)}= \mathbb{E}_{p_{S,X,S}} \left[\log \frac{p\left(m_S^{(l)}|m_{S}^{(l-1)},m_{X}^{(l-1)}\right)}{r\left(m_S^{(l)}\right)} \right]\\
		&=\mathbb{E}_{p_{S,X}}D_{KL}\left(p\left( m_{S}^{(l)}|m_{S}^{(l-1)},m_{X}^{(l-1)}\right)||r\left(m_S^{(l)}\right) \right)\,,\\
	\end{split}
\end{equation}
where $p_{S,X,S}$ represents $p(m_{S}^{(l-1)},m_{X}^{(l-1)},m_S^{(l)})$ and $p_{S,X}$ represents $p(m_{S}^{(l-1)},m_{X}^{(l-1)})$. 
Similarly, we can get the upper bound of $XIB^{(l)}$ by replacing the above variables $m_{S}^{(l-1)}\rightarrow m_{S}^{(l)}$ and $m_{S}^{(l)}\rightarrow m_{X}^{(l)}$, then we have
\begin{equation}
	\label{bound_XIB}
	\begin{split}
		&\widehat{XIB}^{(l)}= \mathbb{E}_{p_{X,S,X} }\left[\log\frac{p\left(m_X^{(l)}|m_X^{(l-1)},m_{S}^{(l)}\right)}{r(m_X^{(l)})}\right]\\
		&=\mathbb{E}_{p_{X,S}}D_{KL}\left(p\left( m_X^{(l)}|m_X^{(l-1)},m_{S}^{(l)}\right)||r\left(m_X^{(l)}\right) \right)\,,\\
	\end{split}
\end{equation}
where $p_{X,S,X}$ represents $p(m_X^{(l-1)},m_{S}^{(l)},m_X^{(l)})$ and $p_{X,S}$ represents $p(m_X^{(l-1)},m_{S}^{(l)})$.
Substituting \eqref{bound_SIB} and \eqref{bound_XIB} into \eqref{lag_mp_GIB} obtains a GIB-based optimization surrogate of communication.

In order to calculate \eqref{bound_SIB} in practice, we need to sample $m_{S}^{(l-1)}$ and $m_{X}^{(l-1)}$ at the $(l-1)$-th layer to calculate the valid structural representation $m_{S}^{(l)}$ at the $l$-th layer. 
After $m_{S}^{(l)}$ is obtained, 
the empirical estimation of $\widehat{SIB}^{(l)}$ is denoted as
\begin{equation}
	\begin{split}
		\overline{SIB}^{(l)}=&\sum_{e\in \mathcal{E}^{(l)}}D_{KL}\left( p_e^{(l)}(m_S^{(l)}|m_{X}^{(l-1)},m_{S}^{(l-1)})||r(m_S^{(l)})\right)\,,
	\end{split}
	\label{es_SIB}
\end{equation}
where $\mathcal{E}^{(l)}$ represents the valid edge set at the $l$-th layer. 
$r(m_S^{(l)})$ is known as the prior structure distribution.
Similarly, we can get the empirical estimation of \eqref{bound_XIB} by replacing the above variables $m_{S}^{(l-1)}\rightarrow m_{S}^{(l)}$ and $m_{S}^{(l)}\rightarrow m_{X}^{(l)}$, then we have
\begin{equation}
	\overline{XIB}^{(l)}=\sum_{v\in \mathcal{V}^{(l)}}D_{KL}\left( p_v^{(l)}(m_X^{(l)}|m_X^{(l-1)},m_{S}^{(l)})||r(m_X^{(l)})\right)\,,
	\label{es_XIB}
\end{equation}
where $\mathcal{V}^{(l)}$ represents the valid node set at the $l$-th layer and $r(m_X^{(l)})$ represents the prior node distribution.

\subsection{Instantiating the GIB principle as CGIBNet} \label{Instantiating}
After the above efforts, we obtain two information-theoretic regularizers based on the GIB principle, \emph{i.e.}, \eqref{es_SIB} and \eqref{es_XIB}. As is shown in Fig. \ref{overall}(a), we instantiate them as a novel communication model, named Communicative Graph Information Bottleneck Network (CGIBNet), which can simultaneously compress the structure and node information to solve the problem of with whom and what to communicate in bandwidth-limited communication.

\textbf{Structure compression learning.}
In order to calculate $\overline{SIB}^{(l)}$ in \eqref{es_SIB} for structural compression,
we first focus on the instantiation of $p^{(l)}_e(\cdot)$. 
Specifically, assuming that $m_{X,i}^{(l-1)} \in \mathbb{R}^{d}$ and $m_{X,j}^{(l-1)} \in \mathbb{R}^{d}$ are the node representations of node $i$ and node $j$ at the $(l-1)$-th layer.
We also suppose the structural representation $m_{S,ij}^{(l)} \in \{0,1\}$ between these two nodes satisfies a Bernoulli distribution.
With these assumptions, as is shown in Fig. \ref{overall}(a.1), we can calculate the parameter $\alpha_{ij}^{(l)}$ of Bernoulli distribution through a link prediction task \cite{zhang2018link}
\begin{flalign}
	&& \qquad\qquad& g_{\alpha}(m_{X,i}^{(l-1)},m_{X,j}^{(l-1)}) && \\
	&& = & {\rm sigmoid}(g_{c}(m_{X,i}^{(l-1)},m_{X,j}^{(l-1)})) && \nonumber\\
	&& = & \alpha_{ij}^{(l)}\,, && \nonumber
\end{flalign}
where the ${\rm sigmoid}$ function constrains the output of the link prediction model $g_c$ to the valid range $(0,1)$.
However, it is difficult to back-propagate the gradients if we directly sample $m_{S,ij}^{(l)}$ from ${\rm Bernoulli}(\alpha_{ij}^{(l)})$. Hence we introduce a relaxed Bernoulli distribution, similar to \cite{jang2017categorical}, to solve it, as
\begin{equation}
	m_{S,ij}^{(l)} \overset{{\rm iid}}{\sim} {\rm RelaxedBernoulli}(\alpha_{ij}^{(l)},\tau)\,,
	\label{bern}
\end{equation}
where $\tau \in (0,\infty)$ is the temperature factor that controls the approximation. 
When $\tau$ is close to 0, $m_{S,ij}^{(l)}$ is approximately sampled from the set $\{0,1\}$. 

In the forward stage, the structural representation $m_S^{(l)}$ at the $l$-th layer can be obtained by repeating the process for all valid edges, as is shown in Fig. \ref{overall}(a.1).
In the compression stage, the prior structure distribution $r(m_S^{(l)})$ in \eqref{es_SIB} is set to ${\rm Bernoulli}(0.5)$ as an uninformative prior.
Then incorporating \eqref{es_SIB} into \eqref{lag_mp_GIB} indicates that the structure representation can be compressed by minimizing the KL divergence between the posterior distribution $p^{(l)}_e(\cdot)$ and the prior distribution $r(m_S^{(l)})$.

\textbf{Node compression learning.}
Following previous IB methods \cite{alemi2017deep,qian2020unsupervised}, we present node embedding as a Gaussian random variable and use the reparameterization trick \cite{kingma2014auto} to tackle the gradient blocking problem. That is, the node message at the $(l-1)$-th layer can be obtained as
\begin{equation} 
	\begin{aligned}
		\epsilon &\overset{{\rm iid}}{\sim} {\rm Gaussian}(0,I) \\ 
		m_{X,v}^{(l-1)} &= \mu_{v}^{(l-1)} \odot \epsilon+\sigma_{v}^{(l-1)} \,,
		\label{gauss}
	\end{aligned}
\end{equation}
where $\odot$ represents the Hadamard product \cite{horn1990hadamard}. $\epsilon$ is sampled from ${\rm Gaussian}(0,I)$.
The mean $\mu_{v}^{(l-1)}\in \mathbb{R}^{d}$ and the variance $\sigma_{v}^{(l-1)}\in \mathbb{R}^{d}$ are obtained by the neural network $g_{\mu,\sigma}$ (Fig. \ref{overall}(a.2)), and $d$ presents the number of bits.
Since $m_S^{(l)}$ has been calculated in the structure learning stage, we can aggregate the information of $m_X^{(l-1)}$ through message passing. Specifically, assuming that node $i$ at the $l$-th layer needs to receive messages from node $j$ and node $k$ according to $m_S^{(l)}$. Then we have
\begin{equation}
	\label{instantiation_node}
	g_{m}(m_{X,i}^{(l-1)},m_{X,j}^{(l-1)},m_{X,k}^{(l-1)})
	=\; m_{X,aggre,i}^{(l)}\,,
\end{equation}
where $m_{X,aggre,i}^{(l)}$ is used as the output of this layer.
In the node compression stage, we set the prior node distribution $r(m_X^{(l)})$ to ${\rm Gaussian}(0,I)$ as is in the previous work \cite{alemi2017deep}.
Then it amounts to minimize the current node information bottleneck of each node message to achieve node compression by \eqref{es_XIB}.
See Algorithm \ref{alg:algorithm} for the pseudo-code of CGIBNet.

\textbf{Drop message bits.} 
Despite that all bits are maintained during the training process, we desire to discard some of them to satisfy the bandwidth constraints.
An important observation is that the node information bottleneck of each message bit represents the amount of task-related information.
Specifically, each bit (or dimension) $b$ of all messages is optimized by
\begin{equation} 
	\min\; \mathcal{L}_{MARL}+\sum\limits_{b} D_{KL}({\rm Gaussian}(\mu_b,\sigma_b)||{\rm Gaussian}(0,I))\,.
\end{equation}
We can find $\mathcal{L}_{MARL}$ and $D_{KL}$ term is a trade-off. Assuming that the current node information bottleneck (\emph{i.e.}, $D_{KL}$ term) of all node bits is 0 after optimization, it means that the KL divergence between the learned Gaussian distribution of all bits and the prior Gaussian distribution ${\rm Gaussian}(0,I)$ is 0, or the learned Gaussian distribution of all bits ${\rm Gaussian}(\mu_b,\sigma_b)$ is equal to the prior Gaussian distribution ${\rm Gaussian}(0,I)$. Since there is no difference between the learned Gaussian distribution of these bits, they do not carry any useful information and $\mathcal{L}_{MARL}$ would not be small in the communication task. We can conclude the following relation based on the above description: if the node information bottleneck (\emph{i.e.}, $D_{KL}$ term) of the message bit is equal to 0, then the learned Gaussian distribution of this message bit is equal to ${\rm Gaussian}(0,I)$ and this message bit is non-informative, and vice versa. Further, if the current node information bottleneck of one particular bit is large while that of the other bits are 0, it means that this bit is constrained by $\mathcal{L}_{MARL}$ and carries most of the task-related information. Therefore, the information amount of each message bit can be measured by the node information bottleneck (\emph{i.e.}, $D_{KL}$ term) and we can rank all bits according to their node information bottleneck, such that bits with a small amount of information are discarded after the training process.
\subsection{Apply CGIBNet to MARL training frameworks}
As discussed in related work, many previous methods (\emph{e.g.}, IC3Net, SchedNet ,IMAC) schedule the structural representations by learning an additional actor so that these methods are not suitable for value-based MARL frameworks.
Instead, we model this task as a link prediction task
to perform end-to-end training without similar restrictions. Therefore, our CGIBNet model can be plugged into both policy-based and value-based MARL frameworks as is shown in Fig. \ref{overall}(b). 
For the value-based framework, 
$\mathcal{L}_{MARL}$ in Equation \eqref{lag_mp_GIB} represents $\mathcal{L}_{MARL}^{QMIX}$, and its calculation follows Equation \eqref{QMIX}.
In this case, ${\rm Agent}_i$ in Fig. \ref{overall}(b) represents each individual $Q_i$ function, and ${\rm Mixing Network}$ $Q_{tot}$ is used to aggregate global information. 
Similarly, for the policy-based framework, 
$\mathcal{L}_{MARL}$ in Equation \eqref{lag_mp_GIB} represents $\mathcal{L}_{MARL}^{MAPPO}$, and its calculation follows Equation \eqref{MAPPO}.
In this case, ${\rm Agent}_i$ in Fig. \ref{overall}(b) represents each individual actor $\pi_i$, and ${\rm Critic Network}$ $V$ is used to aggregate global information. 
	\begin{algorithm}[t]
	\algsetup{linenosize=\footnotesize}
	\small
	\caption{CGIBNet for MARL communication}
	\label{alg:algorithm}
	\SetAlgoLined
	\KwIn{Observation $(o_1,...,o_N)$, edge network $g_{\alpha}$, node network $g_{\mu,\sigma}$, aggregation network $g_{m}$, alive mask $S$, communication round $L$, temperature factor $\tau$}
	\begin{algorithmic}[1] 
		\STATE Initialize the edge set $\{m_{S,jk}=1,1\le j,k \le N\}$, the node 
		
		set $\{m_{X,i}=o_{i},1\le i \le N\}$, and $\overline{SIB}=\overline{XIB}=0$
		\FOR{$l=1,2,\cdots,L$}
		\STATE According to $S$, if any agent $a$ dies, set $m_{X,a}=\emptyset$, $m_{S,a\cdot}=0$, $m_{S,\cdot a}=0$ \,(($\cdot$) $\in$ [$0,N$])
		\FOR{$m_{S,jk}$ in the edge set}
		\IF {$m_{S,jk}=1$}
		\STATE Calculate $\alpha_{jk}=g_{\alpha}(m_{X,j},m_{X,k})$
		\STATE Use $\alpha_{jk}$ to calculate $\overline{SIB}_{jk}$ according to \eqref{es_SIB}
		\STATE Calculate $\overline{SIB}=\overline{SIB}+\overline{SIB}_{jk}$
		\STATE Sample new $m_{S,jk}$ according to \eqref{bern}
		\ENDIF
		\ENDFOR
		
		\FOR{$m_{X,i}$ in the node set}
		\IF {$m_{X,i}\neq \emptyset$}
		\STATE Calculate $\mu_{i},\sigma_{i}=g_{\mu,\sigma}(m_{X,i})$
		\STATE Use $\mu_{i},\sigma_{i}$ to calculate $\overline{XIB}_{i}$ according to \eqref{es_XIB}
		\STATE Calculate $\overline{XIB}=\overline{XIB}+\overline{XIB}_{i}$
		\STATE Sample new $m_{X,i}$ according to \eqref{gauss}
		\ENDIF
		\ENDFOR
		
		\FOR{$m_{X,i}$ in the node set}
		\STATE Concatenate all messages $M={\rm concat}(m_{X,1},\cdots,m_{X,N})$
		\STATE Mask out the corresponding positions that do not send messages to agent $i$ ($m_{S,\cdot i}=0$, ($\cdot$) $\in$ [$0,N$])) in $M$
		\STATE Aggregate messages sent to agent $i$: $m_{X,i}=g_{m}(M)$
		\ENDFOR
		
		\ENDFOR
		
		\STATE Assign $(\boldsymbol{m}_{-1},\boldsymbol{m}_{-2},...,\boldsymbol{m}_{-N})$=
		$(m_{X,1},m_{X,2},\cdots,m_{X,N})$
		\STATE \textbf{return} $(\boldsymbol{m}_{-1},\boldsymbol{m}_{-2},...,\boldsymbol{m}_{-N})$, $\overline{SIB}$, $\overline{XIB}$
	\end{algorithmic}
\end{algorithm}
\section{Experiments}
In this section, we first introduce two metrics to measure bandwidth compression. Then we evaluate CGIBNet in two environments with bandwidth-constrained communication.
The first one is Traffic Control, where we use a policy-based MARL framework (\emph{i.e.,} MAPPO) for training. 
The second is the StarCraft II \cite{samvelyan2019starcraft}, where we use a value-based MARL framework (\emph{i.e.,} QMIX) for training.

\subsection{Evaluation Metric} \label{Evaluation Metric}
\begin{figure*}[htbp]
	\begin{center}
		\includegraphics[width=0.90\linewidth]{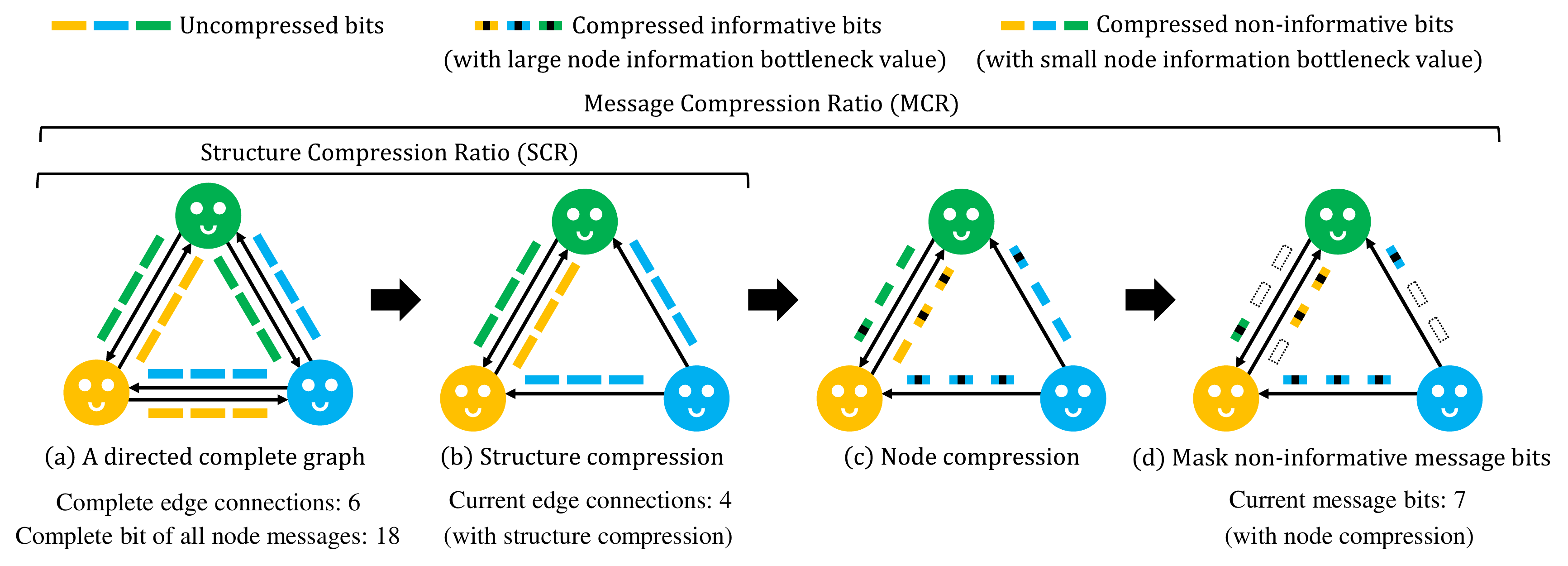}
		\caption{Schematic diagram of calculation of SCR and MCR: $\text{SCR}=(6-4)/6 = 33.33\%$ and $\text{MCR} = (18-7)/18 = 61.11\%$.}
		\label{SCR_MCR}
	\end{center}
\end{figure*}

\textbf{Structure Compression Ratio (SCR).} This metric refers to \cite{kim2019learning}. 
It focuses on the density of edge connections in the communication graph and assumes the message bits transmitted by each edge are consistent.
Formally, 
let $N_{e,complete}$ and $N_{e, compressed}$ denote the number of edges of the communication graph structure without and with structure compression.
Then, SCR is defined as 
\begin{equation}
	\text{SCR}=\frac{N_{e,complete}-N_{e, compressed}}{N_{e, complete}}\times100\% \,.
\end{equation}
For example, as is shown in Fig. \ref{SCR_MCR}(b), there are 4 edge connections in the communication graph after structure compression, thus $\text{SCR}=(6-4)/6 = 33.33\%$.

\textbf{Message Compression Ratio (MCR).} 
This metric refers to \cite{NDQ}.
It focuses on the message bits in the communication graph.
The insight behind MCR is that the length of the message sent by each agent should be adaptive and various in different situations, 
and few node message bits can compress sufficient information for effective communication.
However, the agent network requires the input of fixed-length messages. Therefore, we introduce a binary mask for each message bit to simulate controllable limited bandwidth.
Formally, 
let $N_{m,complete}$ denotes the message bits without compression and $N_{m,compressed}$ denotes the unmasked message bits with compression. Then,
MCR is defined as
\begin{equation}
	\text{MCR}=\frac{N_{m,complete}-N_{m,compressed}}{N_{m,complete}}\times100\% \,.
\end{equation}
We can find MCR can cover all cases of SCR according to their definition, thus
MCR can reflect the degree of both structure and node compression.
In practice, the masking strategy is based on the importance of each message bit.
For methods that include node compression (\emph{e.g.,} NDQ, IMAC, CGIBNet),
we sort all message distributions in ascending order according to their current node information bottleneck, and mask them
accordingly. 
\emph{e.g.}, as is shown in Fig. \ref{SCR_MCR}(d), the non-informative bits are masked first when the communication bandwidth is constrained.
In this case, $\text{MCR} = (18-7)/18 = 61.11\%$.
For the attention-based method (\emph{e.g.,} G2A), we mask their message bits according to attention weights. For other methods, we mask their message bits according to their absolute value.

\subsection{Traffic Control}
\begin{figure}[htbp]
	\begin{center}
		\includegraphics[width=0.99\linewidth]{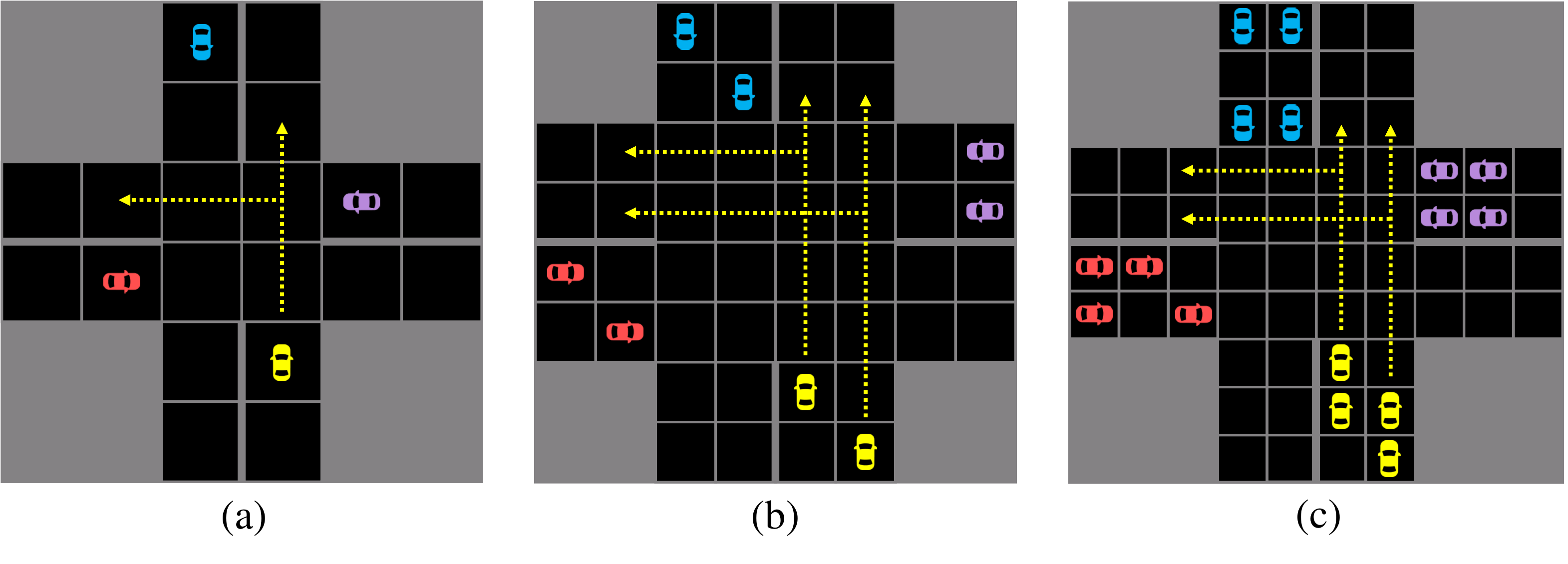}
		\caption{Traffic control environment. (a) easy map (4 cars). (b) medium map (8 cars). (c) hard map (16 cars).}
		\label{TJ_env}
	\end{center}
\end{figure}

\;\;\;\textbf{Environment description.} As is shown in Fig. \ref{TJ_env}, each car is modeled as an agent, and they should learn a good cooperative policy to ensure that they can quickly pass through the junction with as few collisions as possible. The collision only affects the reward of the corresponding agent without changing the environment simulation. 
At the start of the episode, each car needs to randomly initialize a starting position near the entrance and randomly select one of the two possible routes.
The termination condition of the environment is that all cars exit the junction or exceed the maximum timestep (20, 40, 60 for easy, medium and hard maps respectively).
For each car, the local observation contains its location and assigned route id. The action is a discrete number $a \in \{0,1\}$, where $0$ represents stop and $1$ represents move one step forward. The reward consists of three parts: timestep penalty $r_{tp,i}$\;=\;-1, collision penalty $r_{cp,i}$\;=\;-100 and exit bouns $r_{eb,i}$\;=\;30. Thus the shared total reward at timestep $t$ is $r_{tot}=\sum_{i}^{N} (t\times r_{tp,i}+r_{cp,i}+r_{eb,i})$.
The number of grids and cars on each map is consistent with Fig. \ref{TJ_env}(a)-\ref{TJ_env}(c).

\textbf{Experimental setups.} 
In this environment, we modify MAPPO \cite{yu2021surprising} algorithm as the training framework and compare CGIBNet with various baselines, including IC3Net \cite{singh2018learning}, G2A \cite{liu2020multi}, SchedNet \cite{kim2019learning}, ETCNet \cite{ETCNet}, NDQ \cite{NDQ} and IMAC \cite{IMAC},
where IC3Net and G2A only schedule the edge connections in the communication graph while SchedNet and ETCNet explicitly constrains the redundant structural information.
NDQ and IMAC only compress node information and ignore structural information.
Since SchedNet and IMAC learn a scheduler in a similar way and they only consider one of structure compression or node compression, we combine these two methods to construct a strong baseline, which is called SchedIMAC.
For experimental hyperparameters, the training step is 3000(easy)/3000(medium)/10000(hard), the number of bits per message is 5, the discount factor is 0.99, the learning rate for actor/critic
is 0.001/0.003, the batch size is 500, the optimizer is RMSprop. All network backbones refer to MAPPO \cite{yu2021surprising}, where the critic hidden dimension is 32 and the RNN hidden dimension is 64.
The particular hyperparameters of baselines mainly follow their original implementations.
We first set the communication round of all methods to 1 for a fair comparison, because previous methods cannot reduce bandwidth occupancy in multi-round communication.
The structural hyperparameter $\beta_{S}$ and node hyperparameter $\beta_{X}$ of our CGIBNet are set to 0.1 and 0.001 respectively. 

\textbf{Results.} 
Table \ref{TJ_table} shows the average results of 5 experiments on traffic control tasks.
`R' indicates cumulative reward per episode.
The `SCR' related results are obtained by standard evaluation without any restrictions, which indicates the ability of the corresponding model to compress the structure information in the communication graph.
Specifically, SCR=$p$\% in the table indicates that the available communication channels are compressed by $p$\%.
The `MCR' related results are obtained by limiting available message bits during the evaluation process, which ensures that the message compression strength of all methods is consistent so as to make a fair comparison.
Specifically, MCR=$p$\% in the table indicates that the available message bits are compressed by $p$\%.

According to the result of structural compression,
NDQ is the only method to communicate using a fully connected graph (\emph{i.e.}, SCR=0\%) and it achieves poor performance on complex maps, which indicates traffic control tasks rely on scheduling with whom to communicate, otherwise too much redundant information will lead to multi-agent miscoordination.
Interestingly, we find that the performance of CGIBNet is better than IC3Net even though CGIBNet has a larger SCR, which indicates our structural regularizer can effectively cut redundant edge connections in the communication graph.
Additionally, ETCNet, SchedIMAC and CGIBNet are all explicitly structural compression methods with similar SCR on medium and hard maps, but the performance of CGIBNet is much better than that of ETCNet and SchedIMAC, which demonstrates the effectiveness of our method.
Meanwhile, the suboptimal performance of SchedIMAC also verifies that combining existing methods (\emph{i.e.}, SchedNet and IMAC) without a unified theoretical guide is not a good choice.
Although G2A achieves the best performance thanks to its attention mechanism, it needs to occupy more bandwidth than CGIBNet.

According to the result of message compression,
the performance of IC3Net, G2A and ETCNet has significantly reduced under MCR=50\% and MCR=75\% settings.
This is because these methods do not compress the node information, thus a small number of bits cannot express sufficient content for communication. 
The performance of CGIBNet surpasses all other methods under MCR=50\% and MCR=75\% while maintaining a tolerable degradation compared to the standard evaluation, especially for the hard map. 
This indicates that CGIBNet can effectively determine what messages to send and whom to address them so that just a few message bits are enough for communication.
In addition, the performance of each method decreases drastically when all communication channels are closed (\emph{i.e.}, MCR=100\%), proving that the success of our method comes from the compressive communication graph instead of implicit coordination strategies are learned.
\begin{table}[t]
	\tiny
	\centering
	\caption{Experimental results in different restrictions \\on traffic control.}
	\label{TJ_table}
	\begin{tabular}{|c|cc|ccc|}
		\hline
		Map             & \multicolumn{5}{c|}{Easy}                       \\ \hline
		Evaluation Mode & SCR & \tabincell{c}{R under \\the left SCR} & \tabincell{c}{R under \\ MCR=50\%}       & \tabincell{c}{R under\\ MCR=75\%}       & \tabincell{c}{R under\\ MCR=100\%}       \\ \hline
		MAPPO+IC3Net       & 10\%  & 12.5        & -38.3   & -103.4  & -205.5  \\
		MAPPO+G2A          & 7\%  & \textbf{16.2}        & -25.7  & -94.0  & -196.4  \\
		MAPPO+ETCNet          & 16\%  & 12.5        & -20.4  & -62.3  & -236.4  \\
		MAPPO+NDQ          & 0\%   & 13.0        & 3.1    & 0.6    & -289.8  \\
		MAPPO+SchedIMAC         & 0\%   & 12.1        & 4.5    & 1.5    & -277.4  \\
		MAPPO+CGIBNet      & \textbf{17\%}  & 14.8        & \textbf{6.1}    & \textbf{3.9}    & -269.2  \\ \hline
		Map             & \multicolumn{5}{c|}{Medium}                     \\ \hline
		Evaluation Mode & SCR & \tabincell{c}{R under \\the left SCR} & \tabincell{c}{R under \\ MCR=50\%}       & \tabincell{c}{R under\\ MCR=75\%}       & \tabincell{c}{R under\\ MCR=100\%}       \\ \hline
		MAPPO+IC3Net       & 21\%  & 14.6        & -53.5   & -298.9  & -531.2  \\
		MAPPO+G2A          & 18\%  & \textbf{24.1}        & -40.1  & -244.2  & -549.3  \\
		MAPPO+ETCNet          & \textbf{30\%}  & 15.7        & -42.5  & -154.7  & -554.7  \\
		MAPPO+NDQ          & 0\%   & 2.8        & -9.8    & -18.2   & -666.0  \\
		MAPPO+SchedIMAC         & 29\%  & 14.2        & 3.5     & -2.6    & -604.7  \\
		MAPPO+CGIBNet      & \textbf{30\%}  & 19.4        & \textbf{13.2}    & \textbf{9.9}    & -642.1  \\ \hline
		Map             & \multicolumn{5}{c|}{Hard}                       \\ \hline
		Evaluation Mode & SCR & \tabincell{c}{R under \\the left SCR} & \tabincell{c}{R under \\ MCR=50\%}       & \tabincell{c}{R under\\ MCR=75\%}       & \tabincell{c}{R under\\ MCR=100\%}       \\ \hline
		MAPPO+IC3Net       & 20\%  & -1254.0     & -2064.1 & -2946.3 & -4638.3 \\
		MAPPO+G2A          & 19\%  & \textbf{-1108.3}     & -1943.0 & -2857.4 & -3835.1 \\
		MAPPO+ETCNet          & 27\%  & -1387.9        & -1744.3& -2242.0  & -4842.5  \\
		MAPPO+NDQ          & 0\%   & -2529.6     & -2684.2 & -2871.6 & -5934.0 \\
		MAPPO+SchedIMAC         & 27\%  & -1484.4     & -1556.9 & -1773.6 & -5421.4 \\
		MAPPO+CGIBNet      & \textbf{28\%}  & -1195.8     & \textbf{-1231.6} & \textbf{-1289.0} & -6284.7 \\ \hline
	\end{tabular}
\end{table}
\begin{figure}[t]
	\begin{center}
		\includegraphics[width=0.99\linewidth]{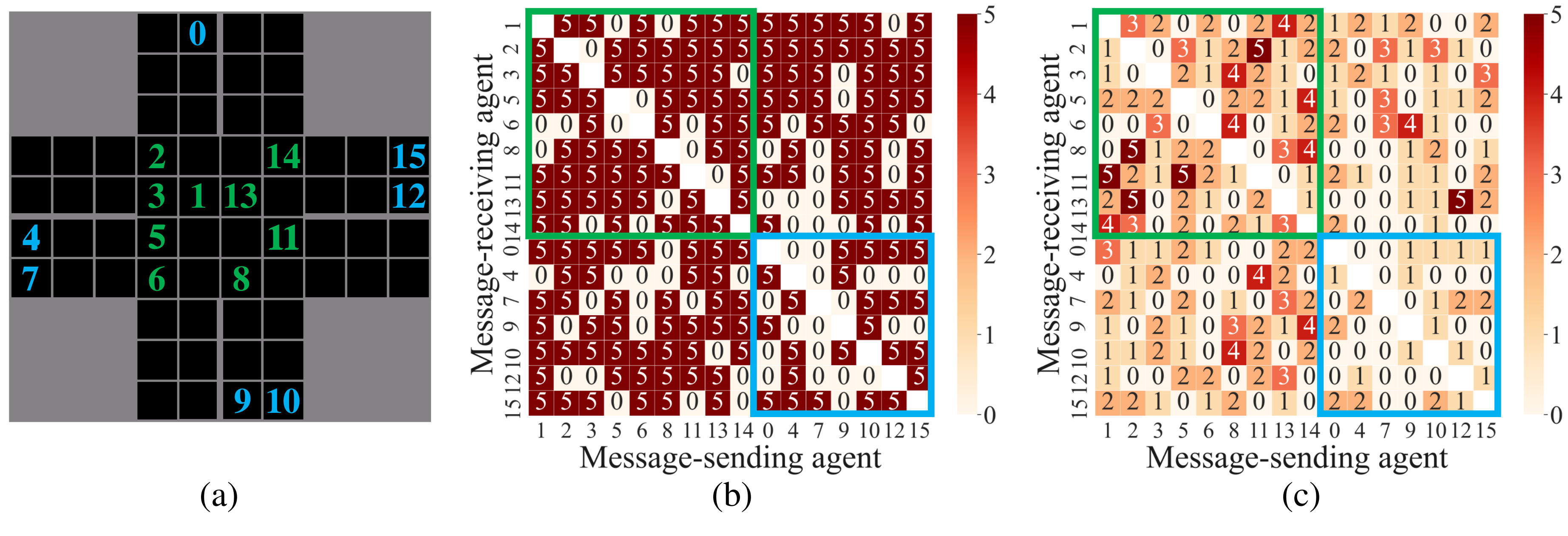}
		\caption{Visualize the message transmission process. (a) Environment state. (b) Confusion matrix under SCR=28\% (message bit is 5). (c) Confusion matrix under MCR=75\%.}
		\label{TJ_vis}
	\end{center}
\end{figure}

\textbf{Visualize the message transmission process.} 
As is shown in Fig. \ref{TJ_vis}, we visualize the communication confusion matrices and node information bottleneck distribution at a certain timestep on the hard map, where the values in Fig. \ref{TJ_vis}(b) and Fig. \ref{TJ_vis}(c) represent the number of message bits. Specifically, Fig. \ref{TJ_vis}(a) illustrates the environment state at one timestep. We can find the agents can be divided into two groups. One group is gathered at the intersection and highlighted in green, and another group is located at the entrance and highlighted in blue.
Fig. \ref{TJ_vis}(b) shows the communication confusion matrix when only structure compression is considered (\emph{i.e.}, SCR=28\%), where
the X-axis represents the ID of the message-sending agent, the Y-axis represents the ID of the message-receiving agent, and the value in the matrix represents the number of message bits.
Fig. \ref{TJ_vis}(c) is similar to Fig. \ref{TJ_vis}(b), but the total message bits are compressed to 75\% (\emph{i.e.}, MCR=75\%). 
From the results, we can find that a lot of message bits are transmitted between the agents at the intersection because they need to prevent collisions at this time (\emph{i.e.}, the green box in Fig. \ref{TJ_vis}(b) and Fig. \ref{TJ_vis}(c)). On the contrary, the agents near the entrance are far apart and therefore communicate less with each other (\emph{i.e.}, the blue box in Fig. \ref{TJ_vis}(b) and Fig. \ref{TJ_vis}(c)).
The above shows that CGIBNet can effectively allocate communication resources in the multi-agent system.
\begin{figure}[t]
	\begin{center}
		\includegraphics[width=0.99\linewidth]{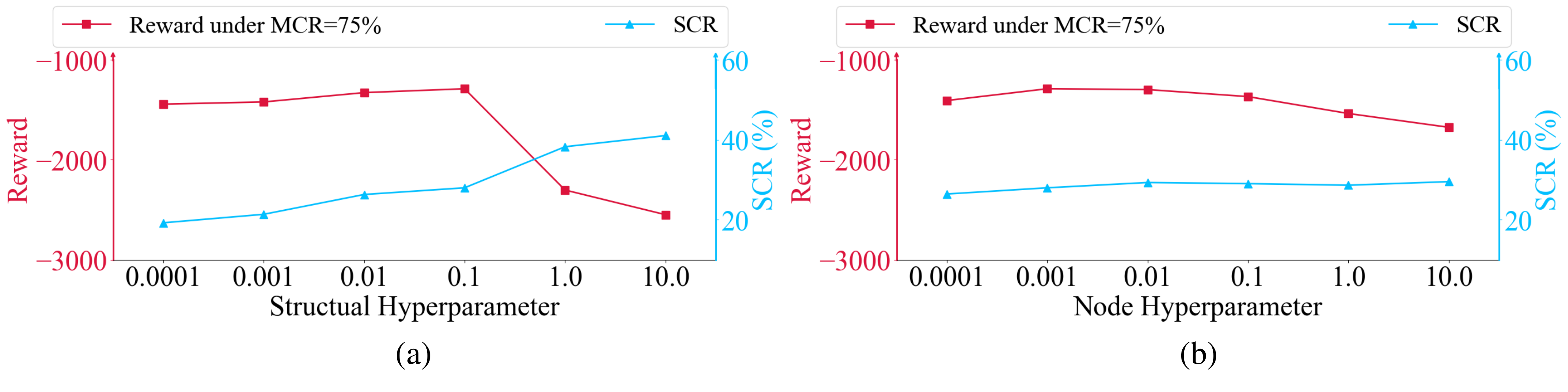}
		\caption{The results on the hard map w.r.t. different hyperparameters $\beta_{S}$ and $\beta_{X}$. (a) $\beta_{S}$ ($\beta_{X}=0.001$). (b) $\beta_{X}$ ($\beta_{S}=0.1$).}
		\label{hyperparameter_choices}
	\end{center}
\end{figure}
\begin{table}[t]
	\tiny
	\centering
	\caption{CGIBNet on hard map w.r.t. different communication round. 22\%/45\% represents the SCR of the 1st and 2nd rounds. MCR focuses on the last round.}
	\label{multi_step}
	\begin{tabular}{|c|cc|ccc|}
		\hline
		\tabincell{c}{Comm. \\round} & SCR & \tabincell{c}{R under \\the left SCR} & \tabincell{c}{R under \\ MCR=50\%}       & \tabincell{c}{R under\\ MCR=75\%}       & \tabincell{c}{R under\\ MCR=100\%} \\ \hline
		1        & 28\%        & -1195.8         & -1231.6 & -1289.0 & -6284.7 \\ \hline
		2        & 22\%/45\%   & -1073.2         & -1092.1 & -1139.6 & -6586.5  \\ \hline
	\end{tabular}
\end{table}

\textbf{The influence of $\bm{\beta_{S}}$ and $\bm{\beta_{X}}$ on CGIBNet.} 
From the optimization perspective, hyperparameters $\beta_{S}$ and $\beta_{X}$ in \eqref{lag_mp_GIB} control the structure and node compression strength respectively. Fig. \ref{hyperparameter_choices} shows that as the compression strength increases (\emph{i.e.}, $\beta_{S}$ and $\beta_{X}$ become larger), the performance of our model under MCR=75\% will gradually rise. However, when they exceed a certain critical value, the information distortion caused by over-compression will lead to performance degradation or even collapse.
\emph{e.g.}, $\beta_{S}$=1.0 in Fig. \ref{hyperparameter_choices}(a).
Therefore, we recommend setting $\beta_{S}$ to $10^{-2}$$\sim$$10^{-1}$ and $\beta_{X}$ to $10^{-3}$$\sim$$10^{-2}$.
Meanwhile, the monotonic relationship between $\beta_{S}$ and SCR in Fig. \ref{hyperparameter_choices}(a) further proves that $\beta_{S}$ can effectively adjust the degree of structural compression in the communication graph.

\textbf{Evaluate the multi-round communication.} 
As discussed in Section \ref{related work}, all previous work about bandwidth-constrained communication cannot be extended to multiple rounds of communication, so we only evaluate our CGIBNet under this setting. 
Also, similar to some non-bandwidth limited communication methods \cite{sukhbaatar2016learning,das2019tarmac}, we mainly focus on two-round communication.
Three or more rounds of communication will lead to the training collapse, and we check there is no MARL communication literature that reports such settings, which may be caused by the oversmoothness problem in the message passing.
As is shown in Table \ref{multi_step}, our CGIBNet can obtain a multi-layer sparse communication graph through structure compression to reduce bandwidth occupation in each communication round. Specifically, in the two rounds of communication, the communication channels of the first and second rounds are compressed to 22\% and 45\%, respectively. 
Based on this sparse graph, we can further discard the message bits with less information in the last communication round to save bandwidth. \emph{i.e.}, the result under MCR=50\% and MCR=75\%. 
More importantly, the performance of two-round communication is better than the single-round counterpart, showing that our method can work well in the complex communication protocol.
\begin{table}[t]
	\tiny
	\centering
	\caption{Experimental results in different restrictions \\on StarCraft II.}
	\begin{tabular}{|c|cc|ccc|}
		\hline
		Map             & \multicolumn{5}{c|}{3b\_vs\_1h1m}               \\ \hline
		Evaluation Mode & SCR & \tabincell{c}{WR under \\the left SCR} & \tabincell{c}{WR under \\ MCR=50\%}       & \tabincell{c}{WR under\\ MCR=75\%}       & \tabincell{c}{WR under\\ MCR=100\%}   \\ \hline
		QMIX+G2A        & 12\%  & 74.9\%         & 51.2\% & 31.9\% & 11.2\% \\
		QMIX+NDQ        & 0\%   & \textbf{78.9\%}         & 73.2\% & 69.0\% & 2.1\%  \\
		QMIX+IMAC$^{*}$       & 0\%   & 77.2\%         & 72.9\% & 67.9\% & 1.8\%  \\
		QMIX+CGIBNet    & \textbf{20\%}  & 78.0\%         & \textbf{74.5\%} & \textbf{69.4\%} & 3.2\%  \\ \hline
		Map             & \multicolumn{5}{c|}{1o2r\_vs\_4r}             \\ \hline
		Evaluation Mode & SCR & \tabincell{c}{WR under \\the left SCR} & \tabincell{c}{WR under \\ MCR=50\%}       & \tabincell{c}{WR under\\ MCR=75\%}       & \tabincell{c}{WR under\\ MCR=100\%}   \\ \hline
		QMIX+G2A        & 15\%  & \textbf{82.7\%}         & 46.4\% & 38.3\% & 18.1\% \\
		QMIX+NDQ        & 0\%   & 81.9\%         & 75.8\% & 72.1\% & 25.2\% \\
		QMIX+IMAC$^{*}$       & 0\%   & 81.6\%         & 74.3\% & 71.2\% & 20.6\% \\
		QMIX+CGIBNet    & \textbf{26\%}  & 82.2\%         & \textbf{78.3\%} & \textbf{76.6\%} & 16.2\% \\ \hline
		Map             & \multicolumn{5}{c|}{5z\_vs\_1ul}               \\ \hline
		Evaluation Mode & SCR & \tabincell{c}{WR under \\the left SCR} & \tabincell{c}{WR under \\ MCR=50\%}       & \tabincell{c}{WR under\\ MCR=75\%}       & \tabincell{c}{WR under\\ MCR=100\%}   \\ \hline
		QMIX+G2A        & 21\%  & 63.6\%         & 31.2\% & 20.6\% & 23.8\% \\
		QMIX+NDQ        & 0\%   & 62.4\%         & 59.8\% & 56.4\% & 31.6\% \\
		QMIX+IMAC$^{*}$       & 0\%   & 61.8\%         & 57.6\% & 53.3\% & 37.7\% \\
		QMIX+CGIBNet    & \textbf{37\%}  & \textbf{64.7\%}         & \textbf{63.3\%} & \textbf{62.1\%} & 34.7\% \\ \hline
		Map             & \multicolumn{5}{c|}{1o10b\_vs\_1r}               \\ \hline
		Evaluation Mode & SCR & \tabincell{c}{WR under \\the left SCR} & \tabincell{c}{WR under \\ MCR=50\%}       & \tabincell{c}{WR under\\ MCR=75\%}       & \tabincell{c}{WR under\\ MCR=100\%} \\ \hline
		QMIX+G2A        & 26\%  & \textbf{84.5\%}         & 59.5\% & 39.8\% & 19.6\% \\
		QMIX+NDQ        & 0\%   & 78.8\%         & 74.4\% & 70.5\% & 10.4\% \\
		QMIX+IMAC$^{*}$       & 0\%   & 77.3\%         & 69.8\% & 67.1\% & 7.1\%  \\
		QMIX+CGIBNet    & \textbf{44\%}  & 83.6\%         & \textbf{81.0\%} & \textbf{79.6\%} & 5.5\%  \\ \hline
	\end{tabular}
	\label{starcraft_table}
\end{table}

\subsection{StarCraft II}
\;\;\;\textbf{Environment description.} 
StarCraft II micromanagement \cite{samvelyan2019starcraft} is a real-time strategy game that is widely used in the MARL community \cite{rashid2018qmix}.
Recently, Wang et al. \cite{NDQ} introduces four maps to explore the communication mechanism in value-based MARL frameworks.
Specifically, these maps include 3b\_vs\_1h1m, 1o2r\_vs\_4r, 5z\_vs\_1ul and 1o10b\_vs\_1r.
The number and letter on the left of `vs' indicate the number and type of agents we trained, and the right is the information of built-in agents.
For the 3b\_vs\_1h1m map, 3 Banelings need to reach the designated location and attack the enemy at the same time.
For 1o2r\_vs\_4r and 1o10b\_vs\_1r maps, 1 Overseer needs to continuously provide the enemy’s vision to guide the allied forces by communication.
For the 5z\_vs\_1ul map, 5 Zealots need to cooperate and communicate frequently to defeat a powerful Ultralisk.

\textbf{Experimental setups.}
We choose QMIX \cite{rashid2018qmix}, a well-known value-based framework, for multi-agent training.
Since IC3Net, SchedNet and ETCNet need to learn an additional actor as a scheduler, they are not suitable for extension into value-based frameworks. Therefore, we mainly compare CGIBNet with G2A, NDQ and IMAC, where IMAC discards its actor-based scheduler (\emph{i.e.}, not compatible with QMIX) and maintains the node compression part.
This modified IMAC is denoted as IMAC$^{*}$.
For experimental hyperparameters, the training episode is 30000, the learning rate is 0.001, the batch size is 64, the optimizer is Adam. All network backbones refer to QMIX \cite{rashid2018qmix}, where mixer hidden dimension is 32 and RNN hidden dimension is 64. Other hyperparameters are consistent with traffic control experiments.

\textbf{Results.}
Table \ref{starcraft_table} shows the average results of 5 experiments on StarCraft II, where `WR' indicates the average win rate.
For all maps, the conclusion is similar to that of traffic control.
Specifically, CGIBNet can maximally reduce the communication channels between agents (\emph{i.e.}, large SCR) and preserve sufficient information for decision-making.
Also, it maintains its powerful ability under MCR=50\% and MCR=75\% settings, which verifies the superiority of our method under the value-based MARL.

\begin{figure}[t]
	\begin{center}
		\includegraphics[width=0.99\linewidth]{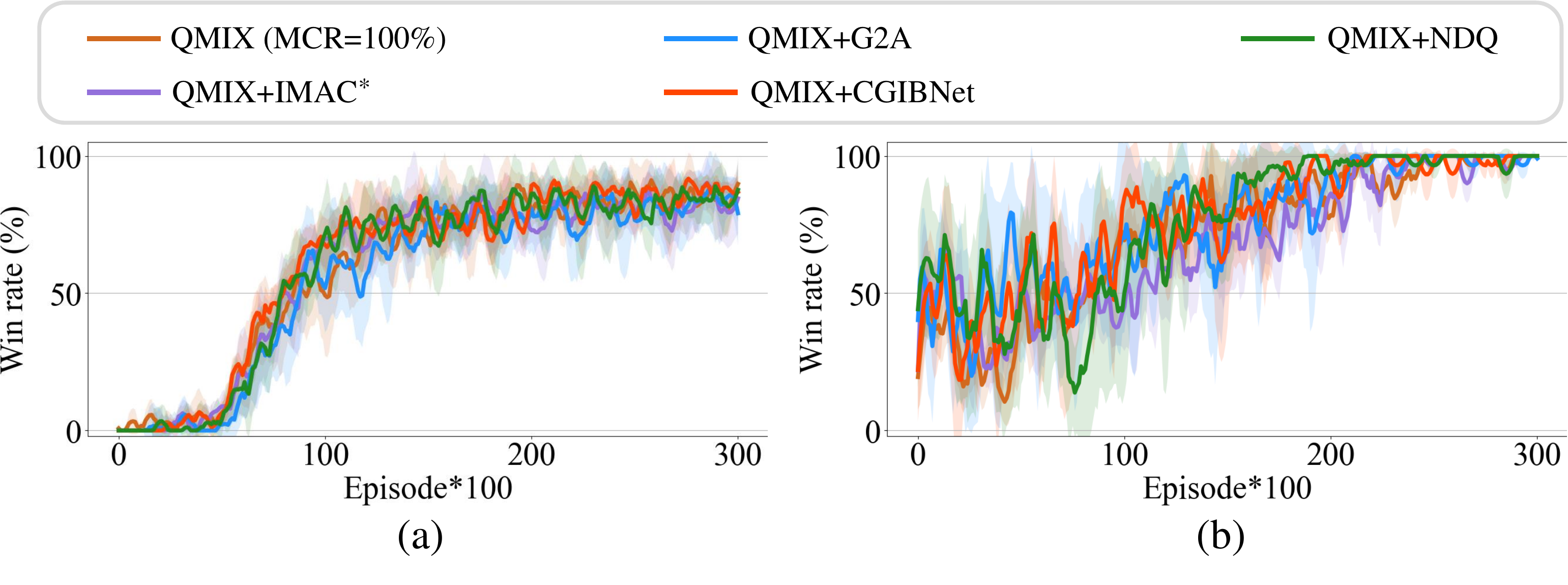}
		\caption{The StarCraft II training curves under MCR=75\%. (a) MMM. (b) bane\_vs\_bane.}
		\label{common map}
	\end{center}
\end{figure}

\textbf{Evaluate on the StarCraft II common maps.}
Samvelyan et al. \cite{samvelyan2019starcraft} propose a number of maps as common benchmarks for MARL algorithms, mainly for the credit assignment problem.
However, we obverse that these maps are not suitable for evaluating communication algorithms, since the sight range of all agents in these maps is sufficient and agents’ position is crowded. This means that each agent can obtain all the information of other agents based on their own private observations without communicating with each other.
We choose MMM map (including 1 Medivac, 2 Marauders \& 7 Marines) and bane\_vs\_bane map (including 20 Zerglings \& 4 Banelings) for verification, respectively.
The result is shown in Fig. \ref{common map}. It can be found that the performance of the algorithm with communication (\emph{i.e.}, G2A, NDQ, IMAC, CGIBNet) and without communication (\emph{i.e.}, vanilla QMIX) is almost the same, which verifies our analysis.

\section{Conclusion}
In this article, we focus on how to learn a communication model to determine each agent is communicating what message and to whom in bandwidth-restricted settings. By modeling multi-agent communication as a directed complete graph, we propose our novel CGIBNet, which leverages the graph information bottleneck principle to compress the flow of node information over the graph structure while maximally preserves information for decision-making. 
We verify that CGIBNet is a universal module, which can be applied to various training frameworks and communication modes.
Empirical results on Traffic Control and StarCraft II show the superiority of CGIBNet under bandwidth-restricted settings.

\bibliographystyle{IEEEtran}
\bibliography{TYCB}

	\vspace{-10 mm}
\begin{IEEEbiography}[{\includegraphics[width=1in,height=1.25in,clip,keepaspectratio]{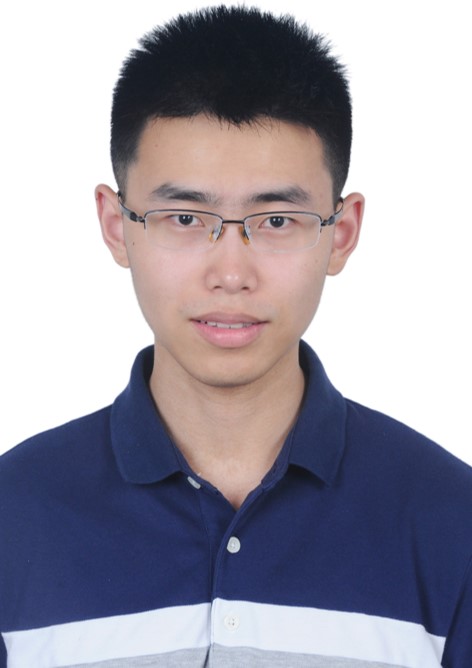}}]{Qi Tian} received the B.S. degree from Chongqing University, Chongqing, China and the M.S. degree from Harbin Institute of Technology, Harbin, China in 2017 and 2019, respectively.
He is currently pursuing the Ph.D. degree with the College of Computer Science and Technology, Zhejiang University, Hangzhou, China, under the supervision of Prof. F. Wu and Assoc. Prof. K. Kuang.
	
His research interests include multi-agent reinforcement learning, machine learning, and adversarial learning.
\end{IEEEbiography}

\begin{IEEEbiography}[{\includegraphics[width=1in,height=1.25in,clip,keepaspectratio]{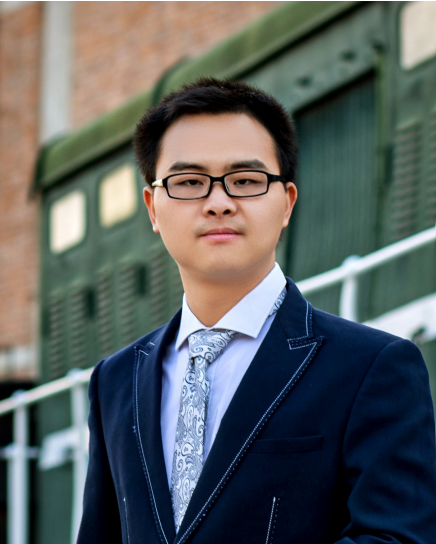}}]{Kun Kuang} received his Ph.D. degree from Tsinghua University in 2019. He is now an Associate Professor in the College of Computer Science and Technology, Zhejiang University. He was a visiting scholar with Prof. Susan Athey’s Group at Stanford University. His main research interests include Causal Inference, Artificial Intelligence, Multi-agent Reinforcement Learning and Causally Regularized Machine Learning. He has published over 40 papers in major international journals and conferences, including SIGKDD, ICML, ACM MM, AAAI, IJCAI, TNNLS, TKDE, TKDD, Engineering, and ICDM, \emph{etc}.
\end{IEEEbiography}	

\begin{IEEEbiography}[{\includegraphics[width=1in,height=1.25in,clip,keepaspectratio]{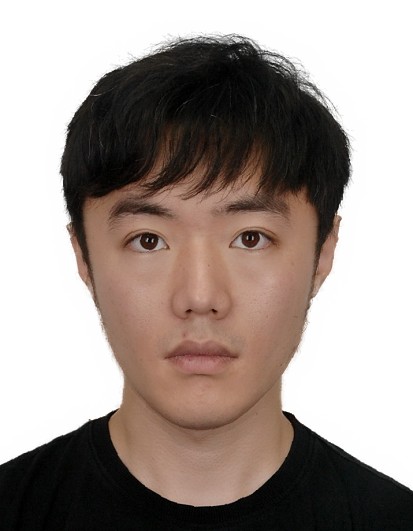}}]{Baoxiang Wang} received the B.E. degree from Shanghai Jiao Tong University, Shanghai, China, in 2014 and the Ph.D. degree from The Chinese University of Hong Kong, Hong Kong, China, in 2020, respectively.
	
He visited University of Alberta and Royal Bank of Canada for 16 months during his Ph.D.. He is currently an assistant professor in The Chinese University of Hong Kong, Shenzhen. The research interests of Baoxiang lie on reinforcement learning, online learning, and learning theory.
\end{IEEEbiography}


\begin{IEEEbiography}[{\includegraphics[width=1in,height=1.25in,clip,keepaspectratio]{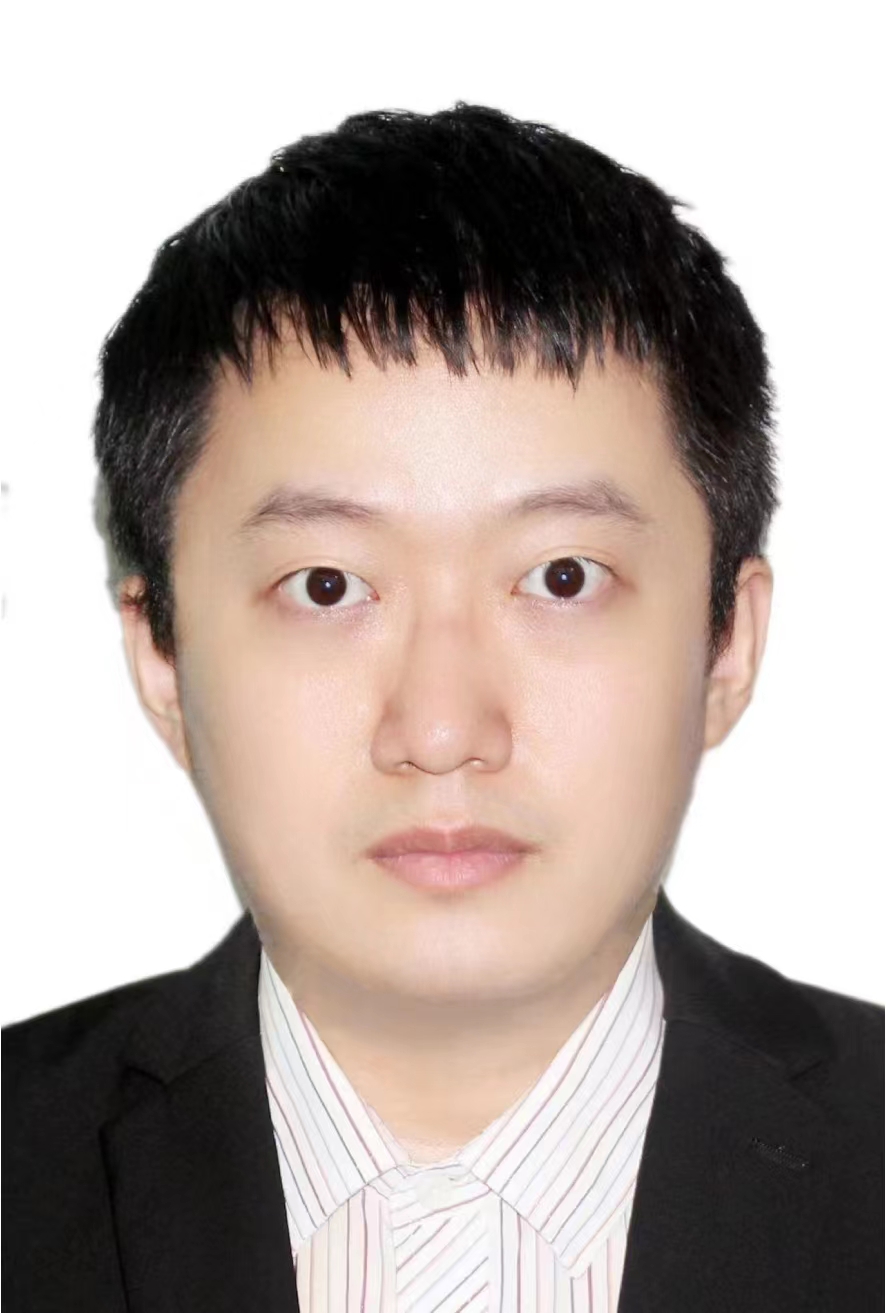}}]{Furui Liu} received the B.S. degree from Zhejiang University, Hangzhou, China, in 2015 and the Ph.D. degree from The Chinese University of Hong Kong, Hong Kong, China, in 2019, respectively.
	
He was a visiting student at Max Planck Institute for Intelligent Systems, and Microsoft Research Asia. 
He is currently a senior researcher at Huawei Noah’s Ark Lab. His research interests include causal inference, deep learning and applications in large scale problems.
\end{IEEEbiography}

\begin{IEEEbiography}[{\includegraphics[width=1in,height=1.25in,clip,keepaspectratio]{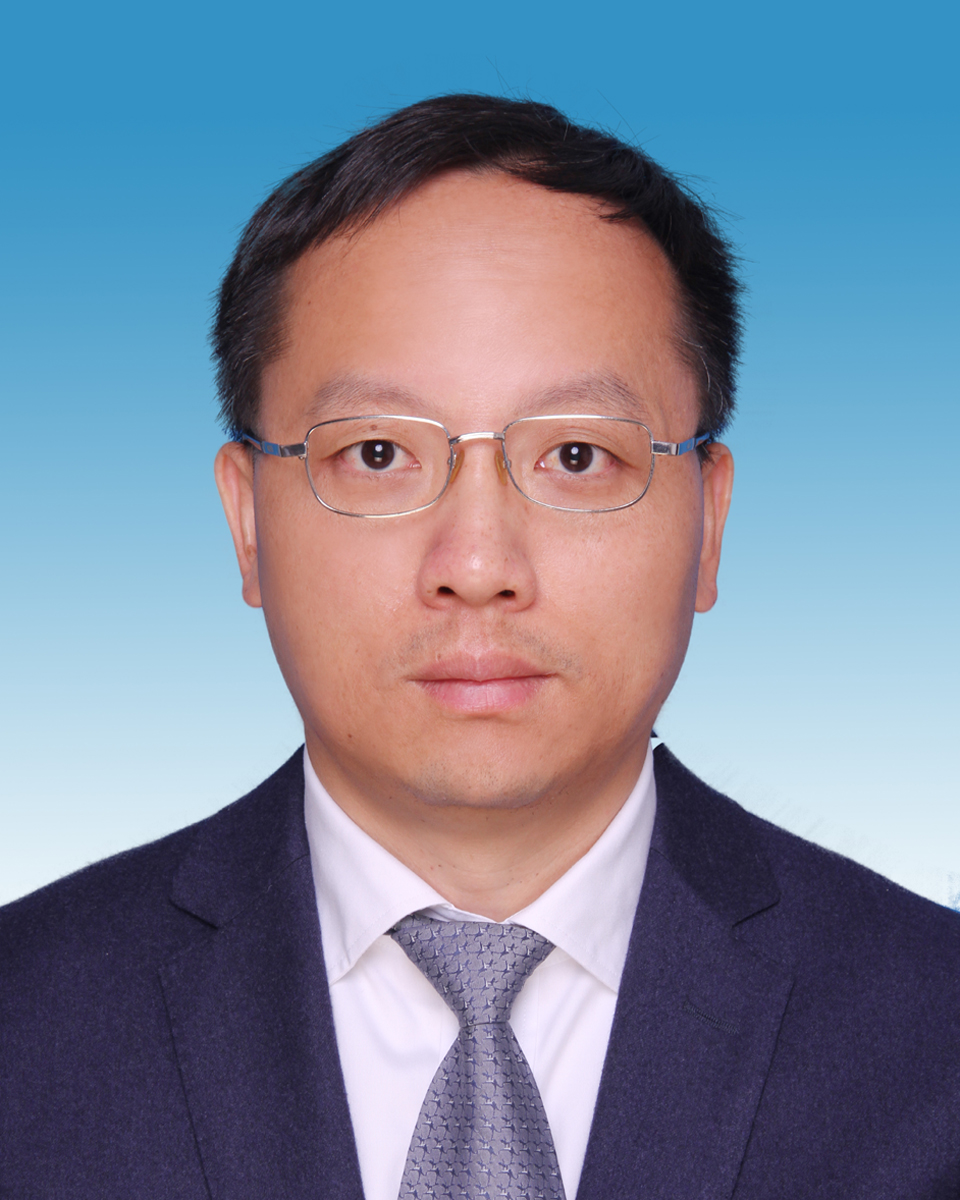}}]{Fei Wu} (Senior Member, IEEE) received the B.S. degree from Lanzhou University, Lanzhou, Gansu, China, the M.S. degree from Macao University, Taipa, Macau, and the Ph.D. degree from Zhejiang University, Hangzhou, China, in 1996, 1999, and 2002.
	
He was a Visiting Scholar with Prof. B. Yu’s Group, University of California at Berkeley, Berkeley, CA, USA, from 2009 to 2010. He is currently a Full Professor with the College of Computer Science and Technology, Zhejiang University. His current research interests include machine learning, sparse representation, and multimedia retrieval.
\end{IEEEbiography}

\end{document}